\documentclass{article}

\usepackage{microtype}
\usepackage{graphicx}
\usepackage{subcaption}
\usepackage{booktabs} 

\usepackage{hyperref}

\usepackage[accepted]{icml2026}

\usepackage{amsmath}
\usepackage{amssymb}
\usepackage{mathtools}
\usepackage{amsthm}

\usepackage{multirow}

\usepackage[capitalize,noabbrev]{cleveref}
\definecolor{lightgray}{gray}{0.92}

\usepackage[table]{xcolor}
\usepackage{caption}
\usepackage{algorithm}
\usepackage{algorithmic}
\usepackage{booktabs}
\usepackage{tikz}
\usetikzlibrary{shapes, arrows.meta, positioning, calc, fit, backgrounds}
\usepackage{tabularx}

\theoremstyle{plain}

\theoremstyle{definition}

\theoremstyle{remark}

\newcommand{\hhy}[1]{\textcolor{black}{#1}}

\usepackage[textsize=tiny]{todonotes}

\icmltitlerunning{Restoring Initial Noise Sensitivity in Text-to-Image Distillation via Geometric Alignment}

\begin{document}

\twocolumn[
  \icmltitle{Restoring Initial Noise Sensitivity in \\ Text-to-Image Distillation  
  via Geometric Alignment
}



  \icmlsetsymbol{equal}{*}
  \icmlsetsymbol{cor}{$\dagger$}

  \begin{icmlauthorlist}
    \icmlauthor{Huayang Huang}{whu}
    \icmlauthor{Ruoyu Wang}{whu}
    \icmlauthor{Jinhui Zhao}{whu}
    \icmlauthor{Wei Deng}{comp}
    \\
    \icmlauthor{Daiguo Zhou}{comp}
    \icmlauthor{Jian Luan}{comp}
    \icmlauthor{Yu Wu}{ai,cor}
    \icmlauthor{Ye Zhu}{sch}
  \end{icmlauthorlist}

  \icmlaffiliation{whu}{School of Computer Science, Wuhan University}
  \icmlaffiliation{comp}{MiLM Plus, Xiaomi Inc.}
  \icmlaffiliation{ai}{School of Artificial Intelligence, Wuhan University}
  \icmlaffiliation{sch}{Laboratoire d’Informatique (LIX), CNRS, École Polytechnique, IPP, France}

  \icmlcorrespondingauthor{Yu Wu}{wuyucs@whu.edu.cn}

  \icmlkeywords{Machine Learning, ICML}

  \vskip 0.3in
]



\printAffiliationsAndNotice{}  

\begin{abstract}
Generative distillation significantly accelerates text-to-image (T2I) generation by compressing multi-step trajectories into few-step student models while preserving perceptual quality. However, existing methods primarily optimize efficiency and output fidelity, often neglecting critical properties of the original trajectory. In this work, we identify a key missing property: sensitivity to initial noise, whose degradation impairs downstream control methods relying on noise-based optimization and manipulation. We trace this issue to standard distillation objectives that enforce pointwise output alignment, inadvertently flattening the input-output landscape and suppressing the teacher’s local geometric structure. To address this, we propose Geometry-Aware Distillation (GAD), a sensitivity-preserving framework that aligns the local functional behavior of teacher and student models. Specifically, GAD matches Jacobian-vector products with respect to input noise, enabling the student to reproduce the teacher’s differential response to perturbations. Extensive experiments across multiple T2I paradigms and noise-driven control tasks demonstrate that GAD significantly restores sensitivity and improves diversity while maintaining high visual fidelity. \hhy{Code is available at \url{https://github.com/Hannah1102/GAD}.}

\end{abstract}

\section{Introduction}
Iterative generative models, encompassing both Diffusion models (DMs) \cite{ho2020denoising,nichol2021glide} and Flow Matching (FM) \cite{lipmanflow,liuflow}, have emerged as the dominant paradigm in modern text-to-image (T2I) synthesis. 
Despite their remarkable generation capabilities, the practical deployment of these models is often hindered by computationally expensive sampling procedures, which typically require tens to hundreds of Neural Function Evaluations (NFEs).
To mitigate this latency, distillation techniques~\cite{meng2023distillation,song2023consistency,sauer2024adversarial} have been rapidly developed. 
By compressing complex multi-step trajectories into single or few-step mappings, these methods successfully reduce inference time by orders of magnitude, making real-time generation increasingly feasible.

\begin{figure}[t!]
    \centering
    \includegraphics[width=\columnwidth]{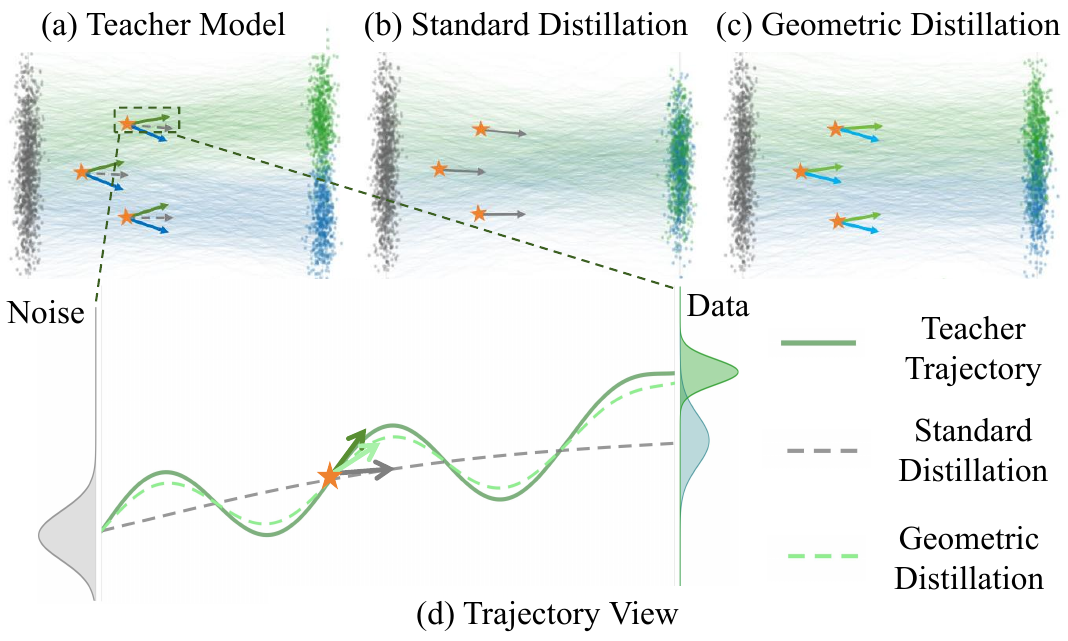}
    \caption{\textbf{Illustration of sensitivity degradation in diffusion distillation.} 
    Top: While the Teacher (a) maps noise to distinct modes (green/blue clusters) with clear directional gradients (arrows), Standard Distillation (b) tends to average these modes, resulting in misaligned gradients. Our Geometry-Aware Distillation (c) successfully recovers the teacher's geometric structure. 
    (d) Trajectory view: standard point-matching (grey dashed) learns a smoothed path with dampened gradients (flattened slope), whereas our GAD (green dashed) preserves the teacher's original curvature and sensitivity to initial noise.}
    \label{fig:intro}
\end{figure}

Existing distillation methods \cite{lu2025adversarial,chen2025nitrofusion} have largely focused on optimizing inference efficiency while maintaining perceptual fidelity.
While successful under standard image quality metrics, this line of work implicitly treats the teacher model as a static input-output mapper and prioritizes alignment of final results.
As a result, several intrinsic properties of the original generative process are often sacrificed, with recent studies highlighting degradation in generation diversity~\cite{gandikota2025distilling}, negative prompt adherence~\cite{nguyen2025supercharged}, and trajectory invertibility~\cite{starodubcev2024invertible}.

In this paper, we identify a critical yet underexplored limitation: \emph{distilled models exhibit severely degraded sensitivity to the initial noise}. 
Rather than being a mere source of randomness, the initial noise acts as a structured prior that determines which specific trajectory the model follows within the vast manifold of images compatible with a given text prompt.
Recent studies suggest that this noise-driven prior provides a complementary axis of control for generative aspects that are difficult to specify via text alone, such as precise spatial layouts~\cite{bancrystal}, low-level visual attributes~\cite{wang2025silent}, and test-time quality enhancement~\cite{eyring2024reno,zhou2025golden}.
More importantly, the creative potential of T2I models relies on the ability to produce diverse visual outputs from the same text prompt, a capability governed by the model's distinct responses to different noise initializations.
However, we observe that compared to the teacher, where different random seeds induce diverse and controllable outputs, distilled models often produce highly correlated samples and respond weakly to noise perturbations. 
This loss of sensitivity not only hampers generative diversity~\cite{gandikota2025distilling,ciderondiversity} but also undermines a broad class of noise-based control techniques~\cite{xie2023boxdiff,zhou2025golden,bancrystal}, whose effectiveness relies on the model's precise responsiveness to input changes.

We attribute this phenomenon to the limitations of the standard distillation objectives.
Specifically, most prior methods formulate distillation as pointwise output alignment, such as minimizing mean squared error \cite{luo2023latent,sauer2024adversarial} or reverse KL divergence \cite{yin2024one,yin2024improved} between teacher and student predictions.
While effective for matching average behavior, these objectives encourage the student to approximate a smoothed conditional expectation over potentially multi-modal outputs \cite{srinivas2018knowledge}. As a result, the local geometric \cite{humayunsecrets} structure of the teacher mapping, particularly its differential response to input perturbations, is suppressed. As illustrated in Fig.~\ref{fig:intro}, this averaging effect flattens the input-output landscape of the student (grey dashed line) and filter out the directional information encoded in the initial noise.

To address this limitation, we propose \textbf{Geometry-Aware Distillation (GAD)}, a sensitivity-preserving framework that aligns local functional behavior between teacher and student models. Instead of solely matching outputs, GAD enforces \emph{response alignment} by constraining the student to replicate the teacher’s Jacobian-vector products with respect to the input. This relational objective preserves local curvature and directional gradients, thus enabling small noise variations to induce meaningful and controllable changes in the synthesized visual outputs.

We conduct a rigorous validation of GAD across diverse generative architectures, spanning SD2 (UNet) \cite{rombach2022high}, PixArt-$\alpha$ (DiT) \cite{chenpixart}, and SANA (flow-based DiT) \cite{xie2025sana}.
We implement GAD within three representative distillation paradigms (output matching \cite{sauer2024fast}, distribution matching \cite{luo2025learning} and score identity distillation \cite{zhou2025score}), benchmarking against 11 distilled baselines. 
Beyond standard metrics, we also employ several initial noise-based control tasks as a rigorous testbed for sensitivity recovery.
Empirical results on layout control and training-free noise retrieval for alignment demonstrate that GAD significantly recovers the effectiveness of noise manipulation compared to standard distillation baselines. Furthermore, we show that restoring sensitivity naturally alleviates diversity degradation without compromising image fidelity, offering a unified solution that reconciles the trade-off between inference speed and generative controllability.

Our contributions are summarized as follows:
\begin{itemize}
    \item We identify the initial noise sensitivity degradation in diffusion distillation and attribute it to the smoothing effect of standard pointwise alignment objectives.

    \item We propose Geometry-Aware Distillation (GAD), a model-agnostic framework that aligns the geometric structure via response alignment, effectively restoring the model's sensitivity to input noise.

    \item We extensively evaluate the controllability and diversity of distilled models, demonstrating that GAD achieves superior performance on multiple downstream tasks dependent on initial noise manipulation.
    
\end{itemize}

\hhy{\textbf{Conflict of Interest Disclosure.} The authors declare that there are no conflicts of interest.}

\section{Background}
\label{sec:background}

\subsection{Generative Models and Initial Noises}
\label{subsec:diffusion_models}

Existing diffusion \citep{ho2020denoising} and Flow Matching models \citep{lipmanflow} can be viewed as learning a velocity field $v_\phi$ that defines a probability flow ODE (PF-ODE) \cite{song2020score}.
Let $\mathbf{x}_t$ denote the state at time $t \in [0, 1]$, where $t=1$ corresponds to the prior distribution $\mathcal{N}(\mathbf{0}, \mathbf{I})$ (pure noise) and $t=0$ corresponds to the real data $\mathbf{x}_0 \sim p_{\text{data}}$ (clean images). The generative process is governed by the ODE:
\begin{equation}
    \frac{d \mathbf{x}_t}{d t} = v_\phi(\mathbf{x}_t, t), \quad \mathbf{x}_1 \sim \mathcal{N}(\mathbf{0}, \mathbf{I}).
    \label{eq:ode}
\end{equation}
In standard diffusion models, $v_\phi$ is derived from the predicted score function $\epsilon_\phi(\mathbf{x}_t, t)$, whereas in Flow Matching, $v_\phi$ is learned directly by regressing target vector fields.
We denote the exact solution of Eq.~\ref{eq:ode} from time $t=1$ to $t=0$ as the \textbf{Teacher Mapping} $\Phi_T: \mathbb{R}^d \to \mathbb{R}^d$.
In practice, sampling from $\Phi_T(\mathbf{x}_1)$ requires $N \in [20, 100]$ function evaluations (NFEs), which creates the latency bottleneck.

Beyond serving as a stochastic starting point, the initial noise $\mathbf{x}_1$ is fundamental to diffusion models.
Under the teacher mapping $\Phi_T$, different initializations induce distinct PF-ODE trajectories, leading to diverse samples even under identical conditioning \cite{sadatcads,um2025minority}.
More importantly, the expressive dependence of $\Phi_T$ on $\mathbf{x}_1$ enables a wide range of noise-based control mechanisms, such as diversity modulation \cite{umboost}, spatial and semantic steering \cite{xie2023boxdiff,li2024enhancing}, and retrieval or manipulation of initial noises \cite{wang2025silent,zhou2025golden} for targeted generation.
This sensitivity to initial noise is a key factor underlying the flexibility of modern diffusion models, as shown in Fig.~\ref{fig:motivation_plots} (c).

\subsection{Diffusion Distillation}
\label{subsec:diffusion_distillation}

Diffusion distillation aims to learn a \textbf{Student Model} $\Phi_S(\cdot; \theta): \mathbb{R}^d \to \mathbb{R}^d$, parameterized by $\theta$, that approximates the teacher's trajectory $\Phi_T$ in a single or few steps.
Given an input noise $\mathbf{z} \sim \mathcal{N}(\mathbf{0}, \mathbf{I})$, the same as the initial state $\mathbf{x}_1$ in the PF-ODE formulation, the student generates a sample $\hat{\mathbf{x}} = \Phi_S(\mathbf{z}; \theta)$.
While specific algorithms differ, most existing distillation methods \cite{kimconsistency,luoyou,lu2025adversarial} can be generalized as minimizing a divergence or distance metric $\mathcal{D}$ between the student's mapping $\Phi_S$ and a target signal $\Phi_T$ derived from the teacher:
\begin{equation}
\label{eq:general_distillation}
\mathcal{L}_{\text{base}} = \mathbb{E}_{\mathbf{z} \sim p(\mathbf{z})} \big[ \mathcal{D} \big( \Phi_S(\mathbf{z}), \Phi_{T}(\mathbf{z}) \big) \big],
\end{equation}
where $\Phi_T(\mathbf{z})$ represents the supervision signal from the teacher (e.g., the ODE trajectory endpoint or a score estimate).
Depending on the specific paradigm, $\mathcal{L}_{\text{base}}$ takes various forms:
\emph{Output Matching} \cite{lin2024sdxl,songimproved} minimizes the discrepancy between student output and teacher trajectory points at certain timesteps, using a regression-based loss (e.g., $L_2$ distance or LPIPS loss).
\emph{Adversarial Alignment} \cite{sauer2024fast,sauer2024adversarial} employs a discriminator to enforce distributional alignment when paired data is scarce or to enhance perceptual quality. 
\emph{Score Matching} \cite{yin2024one,yin2024improved} uses the teacher's score function to provide gradients for the student, avoiding explicit trajectory generation. 
Details about the baseline methods used in our experiments are provided in Appendix~\ref{app:baselines}.

\section{Geometric Gap in Distillation}
\label{sec:motivation}

Standard distillation objectives (Eq.~\ref{eq:general_distillation}) focus on \emph{pointwise alignment}, ensuring that the student $\Phi_S(\mathbf{z})$ matches the teacher $\Phi_T(\mathbf{z})$ for individual inputs.
However, this objective treats inputs $\mathbf{z}$ independently and does not explicitly constrain the \emph{functional landscape} between samples or the local geometry around them. 
This mirrors a known challenge in classical knowledge distillation, where capturing the relational knowledge between samples is often more crucial than mimicking absolute output values \cite{park2019relational, tung2019similarity}.
In the generative context, we hypothesize that this independent treatment overlooks the differential structure of the generative mapping, potentially leading to a misalignment of local gradients. As a consequence, while the student may produce high-fidelity average outputs, it fails to replicate how the teacher responds to input variations-a phenomenon we define as a collapse in \emph{initial noise sensitivity}.

\begin{figure}[t]
    \centering
    \begin{subfigure}[t]{0.47\linewidth}
        \centering
        \includegraphics[width=\linewidth]{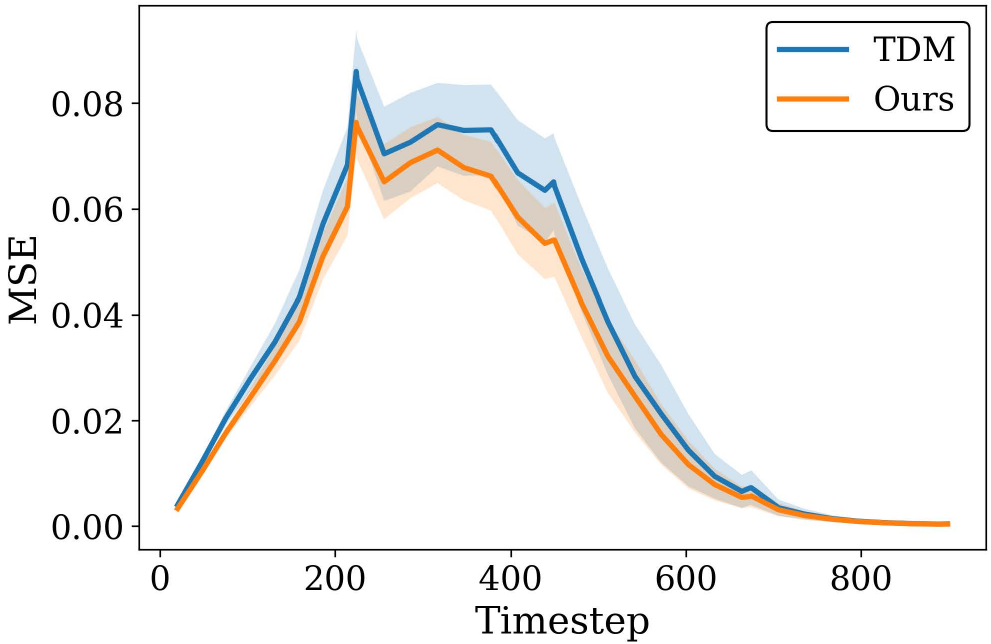} 
        \caption{Pointwise MSE}
        \label{fig:point_jvp-a}
    \end{subfigure}%
    \hfill
    \begin{subfigure}[t]{0.47\linewidth}
        \centering
        \includegraphics[width=\linewidth]{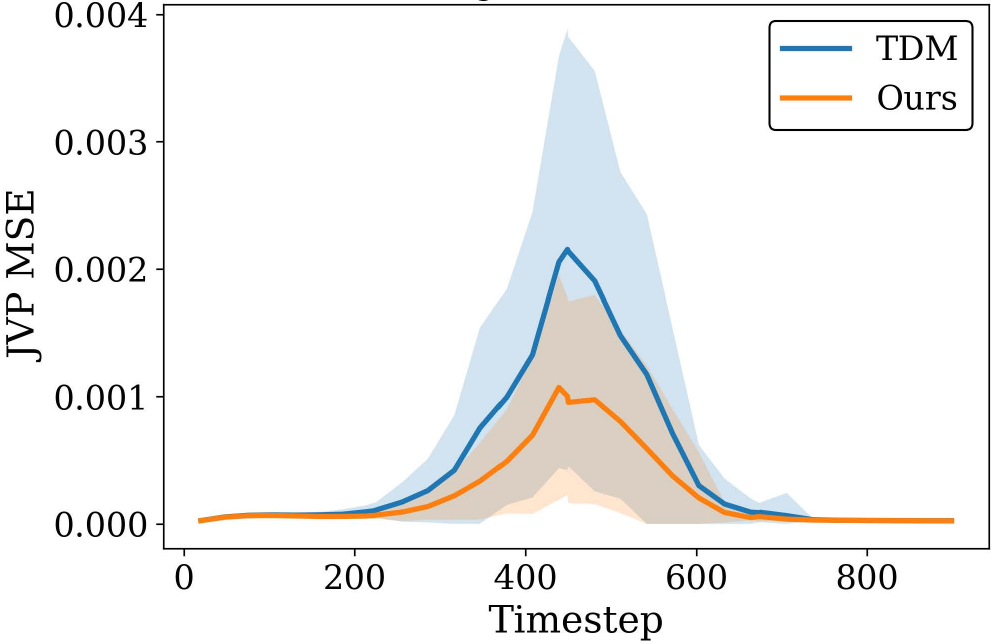} %
        \caption{JVP MSE}
        \label{fig:point_jvp-b}
    \end{subfigure}
     \hfill
     \begin{subfigure}[t]{\linewidth}
        \centering
        \includegraphics[width=\linewidth]{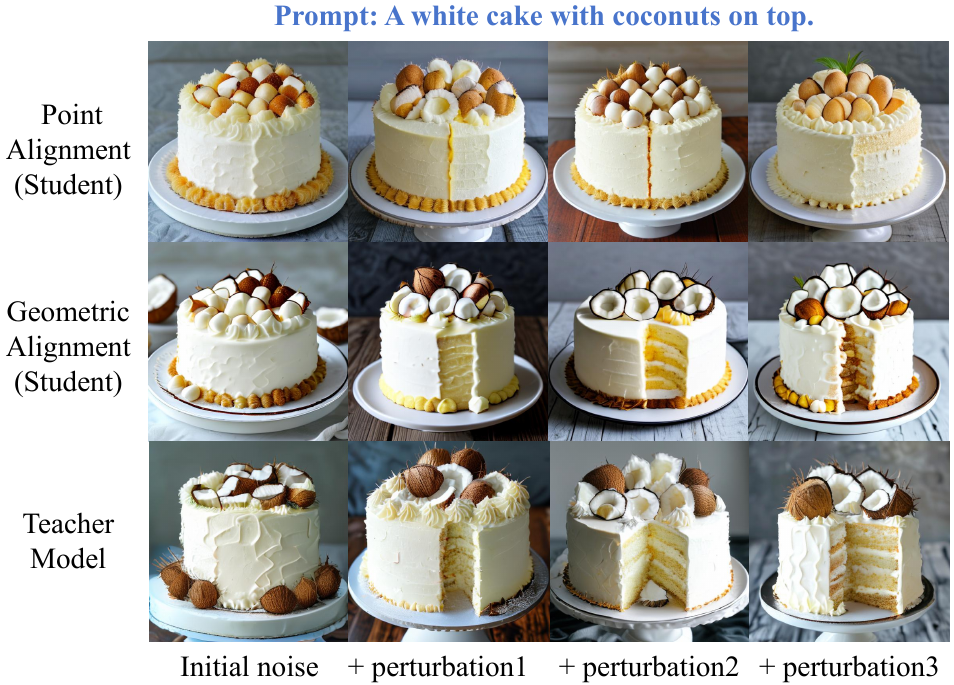} %
        \caption{Visual sensitivity to identical initial-noise perturbations.}
        \label{fig:point_jvp-c}
    \end{subfigure}
    \caption{\textbf{Geometric gap in distillation.} Comparison between baseline TDM (\textcolor{blue}{Blue}) and our method (\textcolor{orange}{Orange}). While the baseline achieves comparable  pointwise MSE to our method \textbf{(a)}, it 
    suffers more from high geometric error \textbf{(b)} and attenuated variations to input perturbations \textbf{(c)}.}
    \label{fig:motivation_plots}
\end{figure}

\textbf{Diagnostic Experiment.}
To empirically verify this hypothesis, we conduct a pilot analysis comparing a standard TDM \cite{luo2025learning} baseline against our GAD (with additional geometric alignment objective). 
We track two metrics after distillation:
\emph{1) Point Alignment:} Measured by Mean Squared Error (MSE) between $\Phi_S(\mathbf{z})$ and $\Phi_T(\mathbf{z})$.
\emph{2) Geometric Alignment:} Measured by the error in Jacobian-Vector Products (JVP), which represents the mismatch in response to input perturbations.
The JVP MSE is computed following the directional derivative objective in Eq.~\ref{eq:nsp_loss}.
Experiments are performed on PixArt-$\alpha$ \cite{chenpixart} using 128 COCO prompts \cite{caesar2018coco}.

\textbf{Observation.}
As shown in Fig.~\ref{fig:motivation_plots}(a), both methods achieve low MSE, indicating that standard distillation is effective at aligning outputs in a \emph{global expectations} sense.
Notably, our geometric constraint does not compromise and even marginally improves point alignment. However, a discrepancy appears in Fig.~\ref{fig:motivation_plots} (b) (JVP Error). The baseline exhibits high error in replicating the teacher's gradient response. It learns to map $\mathbf{z}$ to the correct image, but via a ``smoother" functional path that lacks the teacher's crisp sensitivity.
This manifests visually in Fig.~\ref{fig:motivation_plots}(c), where the point-aligned student exhibits noticeably dampened and less structured changes compared to the teacher.
In contrast, the student trained with our geometric alignment responds in a manner closely matching the teacher, indicating that the local input-output geometry is effectively preserved.

\hhy{\textbf{Direct Measurement of Geometric Alignment.} To provide a more direct assessment of the geometric alignment between teacher and student models, we complement the JVP MSE analysis with additional metrics: JVP Cosine Similarity, Jacobian Norm Ratio, and Spectral KL divergence. These metrics quantify the local differential response and spectral properties of the student relative to the teacher. As summarized in Tab.~\ref{tab:direct_geometry}, incorporating GAD consistently improves alignment across all measures, demonstrating that the student closely replicates the teacher's local geometry.
}

\begin{table}[t]
\centering
\caption{\hhy{\textbf{Direct measurements of geometric alignment.}} Metrics are computed on PixArt-$\alpha$ with baseline TDM.}
\label{tab:direct_geometry}
\small
\setlength{\tabcolsep}{4pt}   
\begin{tabular}{lcccc}
\toprule
\textbf{Method} & \textbf{JVP Cos.}$\uparrow$ & \textbf{Jac. Norm}$\uparrow$ & \textbf{Spec. KL}$\downarrow$ & \textbf{JVP MSE}$\downarrow$ \\
\midrule
\rowcolor{lightgray} Teacher & 1.000 & 1.000 & 0.000 & 0.000 \\
TDM & 0.012 & 0.98 & 0.008 & 0.0003 \\
Ours & \textbf{0.014} & \textbf{0.99} & \textbf{0.006} & \textbf{0.0002} \\
\bottomrule
\end{tabular}%
\vspace{-0.5cm}
\end{table}

\textbf{Conclusion.} These empirical findings reveal that high-fidelity generation (low MSE) does not imply correct underlying dynamics. To bridge this gap, we propose Geometry-Aware Distillation (GAD), which we detail in the following.

\section{Method}
\label{sec:method}

In this section, we introduce Geometry-Aware Distillation (GAD), which serves as a model-agnostic regularization term that can be seamlessly integrated with existing distillation paradigms.
We first present the general theoretical formulation of GAD centered on functional geometry alignment, followed by its specific instantiations within various established distillation frameworks.

\subsection{Geometry-Aware Distillation (GAD)}
\label{subsec:gad}

To restore sensitivity, we propose to explicitly align the local functional behavior of the student $\Phi_S$ with that of the teacher $\Phi_T$. Ideally, we seek to minimize the squared Frobenius norm, which enforces element-wise alignment between the Jacobian matrices:
\begin{equation}
    \mathcal{L}_{\text{Jacobian}}
    = \mathbb{E}_{\mathbf{z} \sim p(\mathbf{z})}
    \left[
        \left\|
            \mathbf{J}_{\Phi_S}(\mathbf{z})
            -
            \mathbf{J}_{\Phi_T}(\mathbf{z})
        \right\|_F^2
    \right],
\end{equation}
where $\mathbf{J}_{\Phi}(\mathbf{z}) = \nabla_{\mathbf{z}} \Phi(\mathbf{z})$.
Following the general formulation in Eq.~\ref{eq:general_distillation}, we let $\Phi(\mathbf{z})$ denote the model's mapping function, which can be instantiated as the predicted trajectory endpoint $\hat{\mathbf{x}}_0$ or the score estimate $\epsilon$ depending on the specific distillation setting.
The input $\mathbf{z}$ is interpreted broadly: it corresponds to the initial Gaussian noise for one-step models, and the intermediate noisy latent state $\mathbf{x}_t$ for few-step distillation. Thus, GAD enforces geometric consistency across the entire sampling trajectory.
However, computing the full Jacobian matrix for high-dimensional image data ($d \approx 10^5$) is computationally prohibitive.

\begin{figure*}[t] %
    \centering
    \includegraphics[width=.85\linewidth]{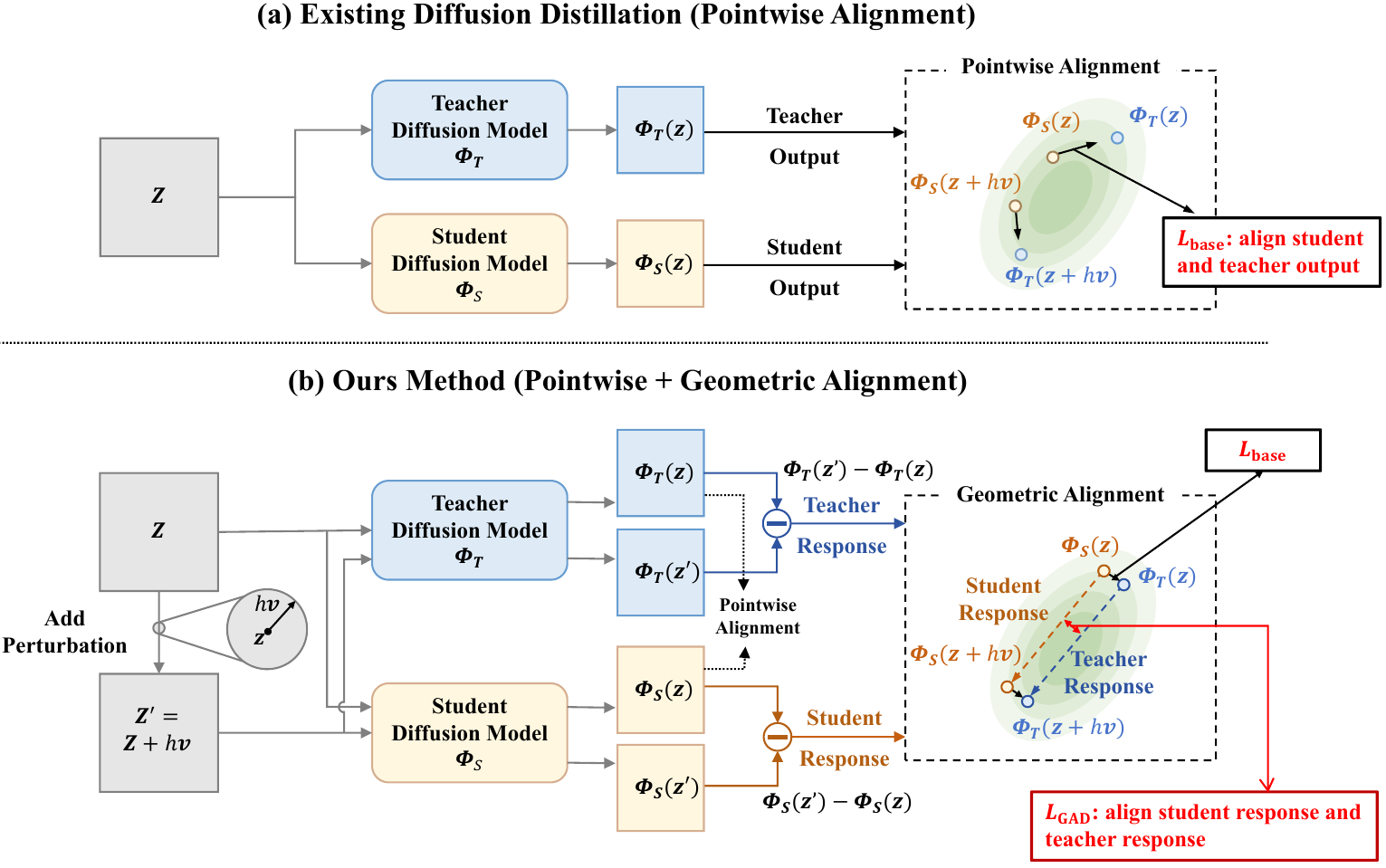} %
    \caption{\textbf{\hhy{Overview of Geometry-Aware Distillation (GAD).}} 
(a) Existing distillation paradigms typically focus on individual pointwise alignment, which often leads the student to learn an ``averaged" direction between $\Phi_{T}(\mathbf{z})$ and $\Phi_{T}(\mathbf{z'})$, thus resulting in a flattened response and loss of diversity.
(b) Our GAD complements the standard loss (dashed) by aligning paired inputs $(\mathbf{z},\mathbf{z'})$ to align the \emph{Response Vectors}.
By explicitly matching the relative change ($\Phi(\mathbf{z}) - \Phi(\mathbf{z}')$), GAD ensures the teacher's output variation is inherited by the student.}
    \label{fig:method_overview}
    \vspace{-0.3cm}
\end{figure*}

\textbf{Jacobian-Vector Product (JVP) Alignment.}
Instead of materializing the full matrix, we propose to align directional derivatives, which capture the essential local geometry. 
For a random perturbation vector $\mathbf{v} \sim \mathcal{N}(\mathbf{0}, \mathbf{I})$, the Jacobian-vector product (JVP)
$\mathbf{J}_{\Phi}(\mathbf{z}) \cdot \mathbf{v}$ \cite{pearlmutter1994fast} characterizes the model's response to the perturbation. We therefore formulate our Geometry-Aware Distillation (GAD) loss as matching these directional responses:
\begin{equation}
    \mathcal{L}_{\text{GAD}}(\theta)
    =
    \mathbb{E}_{\mathbf{z}, \mathbf{v}}
    \left[
        \left\|
            \nabla_{\mathbf{z}} \Phi_S(\mathbf{z}; \theta) \cdot \mathbf{v}
            -
            \nabla_{\mathbf{z}} \Phi_T(\mathbf{z}) \cdot \mathbf{v}
        \right\|_2^2
    \right].
    \label{eq:gad_exact}
\end{equation}
Theoretically, matching the response to random perturbations $\mathbf{v} \sim \mathcal{N}(\mathbf{0}, \mathbf{I})$ is equivalent to minimizing the Frobenius norm of the Jacobian difference in expectation \cite{hutchinson1989stochastic, czarnecki2017sobolev}. Thus, this objective implicitly aligns the full Jacobian geometry.

\textbf{Efficient Approximation via Finite Differences.}
While exact JVP can be computed via forward-mode automatic differentiation \cite{baydin2018automatic,finlay2020train}, it can be memory-intensive or incompatible with certain black-box teacher implementations. To ensure broad applicability and efficiency, we approximate the JVP using a finite difference scheme, as illustrated in Fig.~\ref{fig:method_overview}. 
We perturb the input $\mathbf{z}$ by a small magnitude $h$ in the direction of $\mathbf{v}$:
\begin{equation}
    \nabla_{\mathbf{z}} \Phi(\mathbf{z}) \cdot \mathbf{v}
    \approx
    \frac{
        \Phi(\mathbf{z} + h \mathbf{v}) - \Phi(\mathbf{z})
    }{h},
\end{equation}
Substituting this into Eq.~\ref{eq:gad_exact} and absorbing the constant scaling factor $1/\epsilon^2$ into the loss weight hyperparameter, we arrive at our practical objective:
\begin{equation}
\begin{aligned}
\mathcal{L}_{\text{GAD}}(\theta)
= \mathbb{E}_{\mathbf{z}, \mathbf{v}} \Big[
&\big\|
    \underbrace{\big( \Phi_S(\mathbf{z}') - \Phi_S(\mathbf{z}) \big)}_{\text{Student Response}} \\
&\quad - \underbrace{\text{sg}\big( \Phi_T(\mathbf{z}') - \Phi_T(\mathbf{z}) \big)}_{\text{Teacher Response}}
\big\|_2^2
\Big],
\end{aligned}
\label{eq:nsp_loss}
\end{equation}
where $\mathbf{z}' = \mathbf{z} + h \mathbf{v}$, and $\text{sg}(\cdot)$ denotes the stop-gradient operator to fix the teacher's trajectory. 
Unlike standard distillation (Fig.~\ref{fig:method_overview}(a)), which processes a single input to align pointwise outputs, our method (Fig.~\ref{fig:method_overview}(b)) processes a paired input: the original noise $\mathbf{z}$ and a perturbed version $\mathbf{z}'$.
Both inputs are fed into the teacher and student models. We then compute the \emph{response vectors} by taking the difference between the paired outputs.
The GAD objective strictly enforces that the Student Response ($\Phi_S(\mathbf{z}) - \Phi_S(\mathbf{z}')$) matches the Teacher Response ($\Phi_T(\mathbf{z}) - \Phi_T(\mathbf{z}')$)
This loss forces the student to replicate the \emph{relative change} of the teacher's output given a shift in input, effectively locking the student's local curvature to the teacher's manifold.

\subsection{Unified Training Framework}
\label{subsec:unified_framework}

A distinct advantage of our formulation is that $\mathcal{L}_{\text{GAD}}$ is independent to the pointwise alignment objective. This allows us to apply GAD as a plug-and-play regularizer across diverse distillation paradigms. The total training objective is:
\begin{equation}
    \mathcal{L}_{\text{total}} = \mathcal{L}_{\text{base}} + \lambda \mathcal{L}_{\text{GAD}},
\end{equation}
where $\lambda$ is a balancing hyperparameter. We instantiate this framework across two representative settings:

\textbf{Output Matching:}
In methods like ADD \cite{lin2024sdxl} or LADD \cite{sauer2024fast}, the student $\Phi_S$ directly predicts the trajectory endpoint. Let $\hat{\mathbf{x}}_0 = f_\theta(\mathbf{x}_t, t, c)$ be the model prediction, in which $t$ is the timestep and $c$ is the text condition. The GAD loss ensures that the change in the predicted image mirrors the teacher's response to a perturbed latent $\mathbf{x}_t' = \mathbf{x}_t + h \mathbf{v}$:
\begin{equation}
\mathcal{L}_{\text{GAD}}^{\text{out}} = \mathbb{E}_{\mathbf{x}_t, \mathbf{v}, t, c} \left[ \left\| \Delta \hat{\mathbf{x}}_0^S(\mathbf{x}_t, \mathbf{v}) - \text{sg}\left( \Delta \hat{\mathbf{x}}_0^T(\mathbf{x}_t, \mathbf{v}) \right) \right\|_2^2 \right],
\end{equation}
where $\Delta \hat{\mathbf{x}}_0(\mathbf{x}_t, \mathbf{v}) = f(\mathbf{x}_t + h \mathbf{v}, t, c) - f(\mathbf{x}_t, t, c)$.

\textbf{Score-based Alignment:} For paradigms that align distributions by matching score fields across all timesteps, such as DMD \cite{yin2024one} and SiD \cite{zhou2025few,zhouguided}, the base objective $\mathcal{L}_{\text{base}}$ typically minimizes a divergence between the student-generated distribution $p_{\text{fake}}$ and the real data distribution $p_{\text{real}}$. DMD achieves this by minimizing the KL divergence, while SiD employs the Fisher divergence, enabling a data-free distillation process. In this context, the gradient of the loss involves two score estimators: $\epsilon_\text{real}$, the pre-trained teacher model, and $\epsilon_\text{fake}$, an auxiliary score estimator trained to predict the noise in samples generated by the student. GAD introduces a higher-order geometric constraint by matching the directional variation of these score fields. \hhy{The GAD gradient is formulated as:
\begin{equation}\nabla_{\theta}\mathcal{L}_{\text{GAD}}^{\text{score}} = \mathbb{E}_{\mathbf{x}_t, \mathbf{v}, t, c} \left[  \Delta \epsilon_\text{fake}(\mathbf{x}_t, \mathbf{v}) - \Delta \epsilon_\text{real}(\mathbf{x}_t, \mathbf{v}) \right]\frac{\partial \mathbf{x}_t}{\partial \theta},
\end{equation}
}
where the Score Response $\Delta \epsilon$ is the difference between noise predictions at the original and perturbed locations:
\begin{equation}\Delta \epsilon(\mathbf{x}_t, \mathbf{v}) = \epsilon(\mathbf{x}_t + h \mathbf{v}, t, c) - \epsilon(\mathbf{x}_t, t, c).
\end{equation}
Intuitively, while $\mathcal{L}_{\text{base}}$ aligns the first-order moments (ensuring the student moves toward high-density regions of the teacher's distribution), $\mathcal{L}_{\text{GAD}}^{\text{score}}$ aligns the local curvature and divergence of the score fields. This ensures that the geometric structure of the synthesized manifold consistently follows the teacher's guidance even under local perturbations.
The comprehensive list of hyperparameters and model-specific configurations is included in Appendix~\ref{app:impl_details}.

\section{Experiment}

We first analyze the noise sensitivity of distilled models, followed by an evaluation of the impact of GAD on general generation quality. Finally, we show comparisons on downstream tasks that rely on noise sensitivity for control.

\subsection{Loss of Sensitivity in Distillation}
\label{subsec:exp_sensitivity_analysis}
We first design a diagnostic experiment to empirically verify our core hypothesis that existing diffusion distillation methods suffer from a collapse in \emph{initial noise sensitivity}:
variations in the initial noise no longer induce sufficiently distinguishable changes in the generated images.

\textbf{Experimental Protocol.}
We formulate sensitivity as a \emph{seed identifiability} problem by asking: can a model distinguish between images generated from different initial seeds?
We generate images using 10 fixed seeds with MS-COCO \cite{lin2014microsoft} prompts and train an EfficientFormer-L3 \cite{li2022efficientformer} classifier to predict the source seed.
For each seed, we generate 500 training, 100 validation, and 100 testing images.
We report two metrics:
(1) \emph{Self-Identifiability}: the classifier is trained and tested on the same model to measure the intrinsic noise sensitivity;
and (2) \emph{Teacher Alignment}: the classifier is trained on the Teacher (SD2) and tested on the student models, probing whether the distilled model preserves the semantic trajectory of the teacher.

\begin{table}[t]
  \centering
  \caption{\textbf{Quantitative analysis of seed sensitivity and teacher alignment.}  
  Classification accuracy (\%) of predicting initial seeds from generated images. Self-Identifiability: train/test on self. Teacher Alignment: train on teacher, test on student.}
  \label{tab:seed_sensitivity}
  \resizebox{\linewidth}{!}{
  \begin{tabular}{l|cc}
    \toprule
    \textbf{Distilled Model} & \textbf{Self-Identifiability} & \textbf{Teacher Alignment} \\
    \midrule
    \rowcolor{lightgray}
    SD2 (Teacher) & 93.70\% & - \\
    \midrule
    SD-Turbo \cite{sauer2024adversarial} & 77.80\% & 63.20\% \\
    SwiftBrush \cite{nguyen2024swiftbrush}& 52.90\% & 45.80\% \\
    SwiftBrushv2 \cite{dao2024swiftbrush}&  76.50\%& 67.60\% \\
    TCD \cite{zheng2024trajectory}& 87.30\% & 84.50\% \\
    LADD \cite{sauer2024fast}& 87.60\% & 83.70\% \\
    \textbf{Ours} & \textbf{92.40\%} & \textbf{87.40\%} \\
    \bottomrule
  \end{tabular}
  }
\end{table}

\textbf{Results.} As illustrated in Tab.~\ref{tab:seed_sensitivity}, the Multi-step Teacher exhibits high classification accuracy, indicating that the teacher preserves strong and discriminative variations across different initial seeds.
In contrast, classifiers trained and tested on standard distilled models exhibit significantly lower accuracy.
Our model, which is built upon the LADD with additional geometric sensitivity regularization, restores this sensitivity to 92.40\%, closely matching the teacher.
In terms of Teacher Alignment, our method achieves an accuracy of 87.40\%, outperforming baselines.
This result demonstrates that our GAD not only recovers the magnitude of noise sensitivity but also ensures that the \emph{directional} response to noise perturbations remains consistent with the teacher model.

\begin{figure*}[t]
  \centering
  \includegraphics[width=\linewidth]{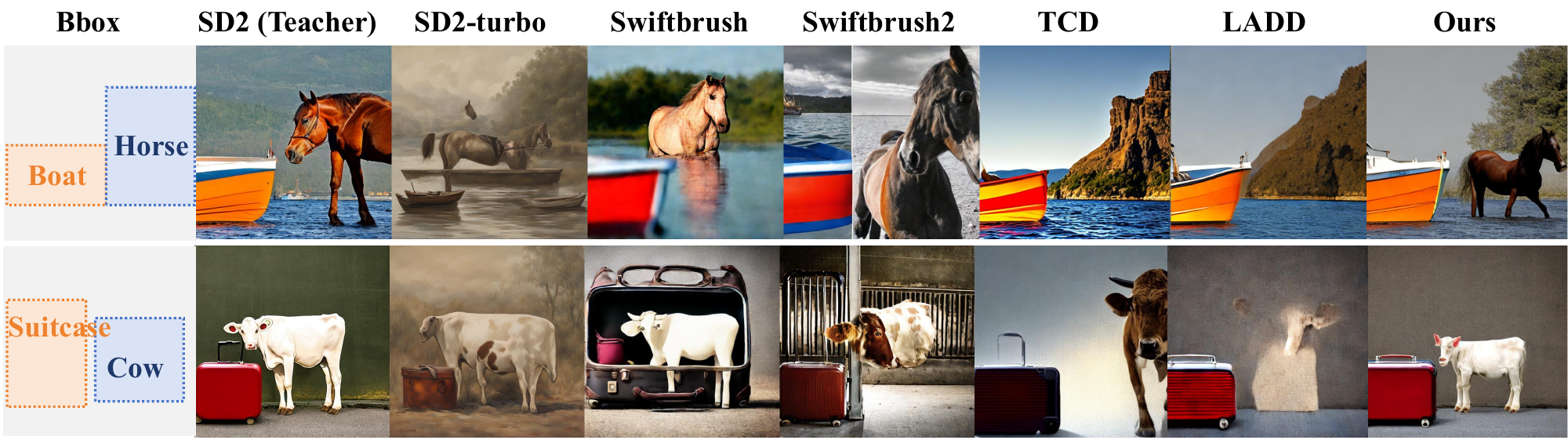}
  \caption{\textbf{Qualitative comparison of layout control.} The left column shows the target bounding boxes. 
  The text prompts are ``A horse and a boat.'' (first row) and ``A cow and a suitcase.'' (second row).}
  \label{fig:layout_control}
\end{figure*}

\subsection{Impact on General Generation Quality}
\label{subsec:general_quality}
 
To ensure geometric regularization does not compromise image fidelity, we evaluate PickScore \cite{kirstain2023pick} and CLIP Score \cite{radford2021learning} on 1000 MS-COCO prompts \cite{lin2014microsoft} to assess image fidelity and text-image alignment.
As reported in Tab.~\ref{tab:general_quality}, our GAD slightly improves general generation quality across all evaluated settings. 
For instance, adding GAD to SiD \cite{zhou2025score} increases the CLIP Score from 32.75 to 34.40.
This consistency proves that our method effectively restores noise sensitivity without disrupting the generative capabilities established by the base distillation methods.
Unlike standalone methods that might be tied to specific architectures, GAD acts as a versatile enhancer that can be seamlessly integrated into existing state-of-the-art pipelines.

\begin{table}[t]
  \centering
  \caption{\textbf{Impact on general generation quality.} 
  Our GAD regularization even slightly improves generation quality.}
  \label{tab:general_quality}
  \resizebox{\linewidth}{!}{
  \begin{tabular}{lcccc}
    \toprule
    \textbf{Method} & \textbf{Architecture} & \textbf{Steps} & \textbf{CLIP} $\uparrow$ &  \textbf{Pickscore} $\uparrow$\\
    \midrule
    \multicolumn{5}{l}{\textit{Setting A: Output Matching with Adversarial Alignment on UNet}} \\
    \rowcolor{lightgray}
    Teacher (SD v2.1) & UNet & 50 & 34.56 & 21.7022 \\
    LADD \cite{sauer2024fast} & UNet & 1 & 32.58 & 20.8515 \\
    \textbf{LADD + Ours} & UNet & 1 & \textbf{32.68} & \textbf{21.3790} \\
    \midrule
    \multicolumn{5}{l}{\textit{Setting B: Distribution Matching on DiT}} \\
    \rowcolor{lightgray}
    Teacher (PixArt-$\alpha$) & DiT & 20 & 33.41 & 22.2922 \\
    TDM \cite{luo2025learning} & DiT & 4 & 33.43 & 22.1007 \\
    \textbf{TDM + Ours} & DiT & 4 & \textbf{33.52} & \textbf{22.2704} \\
    \midrule
    \multicolumn{5}{l}{\textit{Setting C: Score Identity Distillation on Flow-DiT}} \\
    \rowcolor{lightgray}
    Teacher (SANA) & Flow-DiT & 20 & 35.16 & 22.4056 \\ %
    SiD \cite{zhou2025score} & Flow-DiT & 4 & 32.75 & 21.7629 \\ %
    \textbf{SiD + Ours} & Flow-DiT & 4 & \textbf{34.40} & \textbf{22.0735} \\
    \bottomrule
  \end{tabular}
  }
  \vspace{-0.2cm}
\end{table}

\hhy{\textbf{Analysis.} We attribute this enhancement to GAD's better generalization. Unlike standard pointwise distillation that treats inputs independently and risks overfitting to isolated samples, GAD enforces local neighborhood consistency. This yields a smoother, more faithful approximation of the teacher's manifold. Consequently, GAD produces denoising trajectories that are more consistent with the teacher on unseen inputs. As empirically validated in Tab.~\ref{tab:trajectory_drift}, tracking the discrepancy between denoised latents on 200 unseen prompts reveals that GAD consistently reduces cumulative trajectory deviation, achieving a 13\% lower final error. This tighter trajectory alignment preserves teacher-consistent dynamics, explaining the observed fidelity gains.
}

\begin{table}[t]
\centering
\caption{\hhy{\textbf{Average cumulative trajectory deviation from the teacher on PixArt-$\alpha$.}} Lower values indicate that the student's trajectory is more consistent with the teacher.}
\label{tab:trajectory_drift}
\setlength{\tabcolsep}{3pt}
\small
\begin{tabular}{lcccc}
\toprule
\multirow{2}{*}{\textbf{Method}} & \textbf{Stage 1}  & \textbf{Stage 2} & \textbf{Stage 3}  & \textbf{Final} \\
 & ($t=0.75$)$\downarrow$ & ($t=0.5$)$\downarrow$ & ($t=0.25$)$\downarrow$ & ($t=0$)$\downarrow$ \\
\midrule
TDM & 0.016 & 0.216 & 0.433 & 0.491 \\
TDM + Ours & \textbf{0.014} & \textbf{0.184} & \textbf{0.373} & \textbf{0.427} \\
\bottomrule
\end{tabular}%
\vspace{-0.3cm}
\end{table}

\subsection{Downstream task 1: Layout Control}
\label{subsec:layout_control}

We next quantitatively evaluate whether restoring noise sensitivity translates into improved controllability.
We consider a training-free layout control task \cite{xie2023boxdiff}, where the attention-based spatial constraints are injected exclusively through the initial noise.

\textbf{Experimental Protocol.} 
We evaluate performance using 800 prompts from COCO dataset \cite{caesar2018coco} with bounding box annotations from \cite{xie2023boxdiff}.
Layout fidelity is measured via Average Precision (AP) and AP$_{50}$ \cite{li2021image} using YOLOv4 \cite{bochkovskiy2020yolov4}, alongside CLIP score \cite{radford2021learning} for semantic alignment.
We implement our method on top of the LADD framework \cite{sauer2024fast} and compare it against representative distillation baselines.

\textbf{Results.}
Tab.~\ref{tab:layout_control} summarizes the results. The teacher model (SD v2.1) \cite{rombach2022high} establishes an upper bound with an AP of 6.6.
Most distilled baselines exhibit a substantial performance drop (e.g., SD-Turbo drops to 3.0 AP), indicating that spatial information in the initialization is largely washed out.
In contrast, our method achieves the strongest performance among students, significantly outperforming the base LADD model and recovering $87\%$ of the Teacher’s layout accuracy (AP 5.8).
Qualitative results in Fig.~\ref{fig:layout_control} further confirm that while baselines often suffer from object neglect or spatial entanglement, our method faithfully respects the prescribed bounding boxes.
These results demonstrate that preserving geometry effectively prevents the loss of spatial constraints encoded in the initial noise.

\begin{table}[t]
\centering
\caption{\textbf{Quantitative comparison on layout control.} We report AP to measure layout fidelity, and CLIP score for text alignment. 
}
\resizebox{\linewidth}{!}{
\begin{tabular}{lccc c}
\toprule
\textbf{Model} & \textbf{AP} $\uparrow$ & \textbf{AP\textsubscript{50}} $\uparrow$  & \textbf{CLIP} $\uparrow$ \\
\midrule
\rowcolor{lightgray}
SD2 (Teacher Model)     & 6.6 & 21.3 & 0.3333 \\
\midrule
SD-turbo \cite{sauer2024adversarial}& 3.0 & 8.4 & \textbf{0.3237} \\
Swiftbrush \cite{nguyen2024swiftbrush}& 3.8 & 12.6 & 0.3147 \\
Swiftbrushv2 \cite{dao2024swiftbrush}& 2.9 & 10.6 & 0.3203 \\
TCD \cite{zheng2024trajectory}&  4.8 & 18.0 & 0.3169\\
LADD \cite{sauer2024fast}& 5.0 & 17.4 &0.3187 \\
\textbf{Ours} & \textbf{5.8} & \textbf{20.6} & 0.3184 \\
\bottomrule
\end{tabular}
}
\label{tab:layout_control}
\vspace{-0.3cm}
\end{table}

\subsection{Downstream task 2: Generation Diversity}
\label{subsec:diversity_eval}

A common failure mode of distilled diffusion models is conditional mode collapse \cite{yin2024one}, where different random seeds yield nearly identical outputs for a fixed prompt.
We evaluate whether restoring noise sensitivity alleviates this issue.

\textbf{Experimental Protocol.}
Following \cite{kirchhof2025shielded}, we use the CC12M dataset \citep{changpinyo2021conceptual}. For each text prompt, we generate $8$ images using different random seeds.
We employ Vendi Score \cite{dan2023vendi} to measure diversity and CLIP score for text-image alignment.
We incorporate various publicly available distilled models for a comprehensive comparison. For SD v2.1, we evaluate SD-Turbo, SwiftBrush/v2, TCD, and LADD. For PixArt-$\alpha$, we incorporate DMD \cite{yin2024one}, LCM \cite{luo2023latent}, YOSO \cite{luoyou}, FLASH \cite{chadebec2025flash}, and TDM \cite{luo2025learning}. For SANA, we utilize the SiD \cite{zhou2025score} method.
Our method is implemented on top of LADD, TDM and SiD for SD2, PixArt-$\alpha$, and SANA.

\begin{figure}[t] 
    \centering

    \begin{subfigure}[t]{0.32\linewidth}
        \centering
        \includegraphics[width=\linewidth]{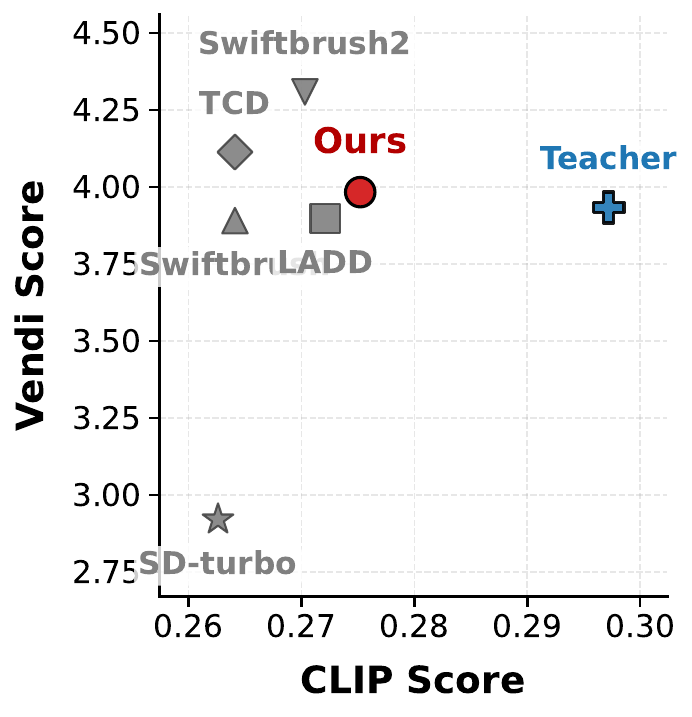}
        \caption{SD2.1}
        \label{fig:div_sd2}
    \end{subfigure}
    \hfill %
    \begin{subfigure}[t]{0.32\linewidth}
        \centering
        \includegraphics[width=\linewidth]{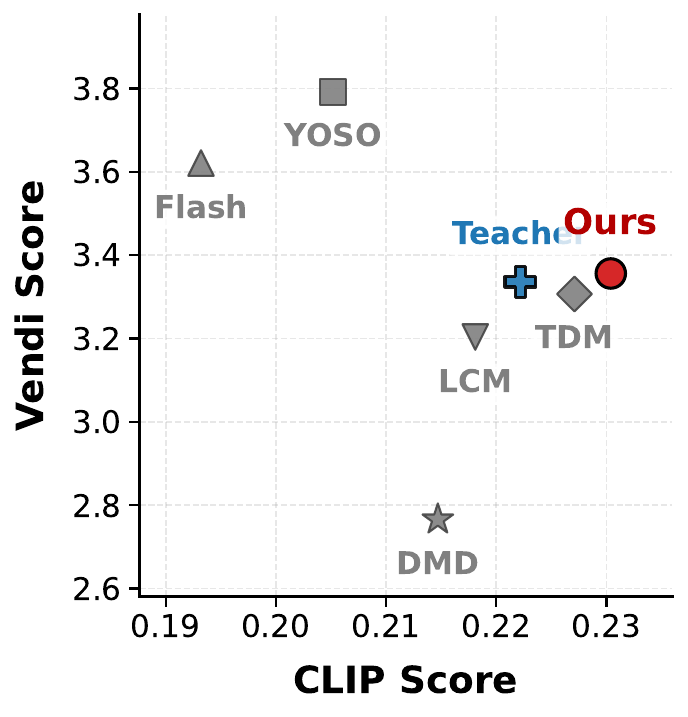}
        \caption{PixArt-$\alpha$}
        \label{fig:div_pixart}
    \end{subfigure}
    \hfill %
    \begin{subfigure}[t]{0.32\linewidth}
        \centering
        \includegraphics[width=\linewidth]{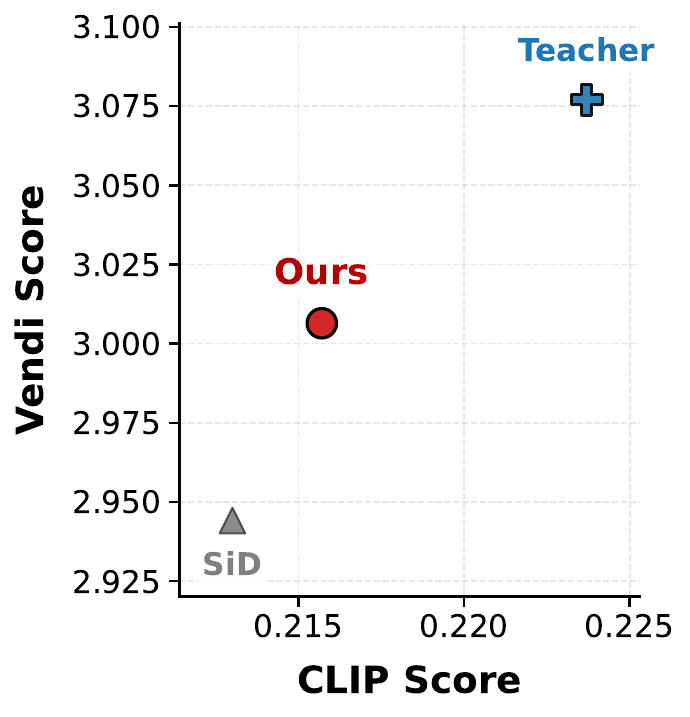}
        \caption{SANA}
        \label{fig:div_sana}
    \end{subfigure}
    \caption{\textbf{Diversity vs. fidelity trade-off.} 
    Vendi Score (Diversity) vs. CLIP Score across three architectures. 
    Baseline methods (grey) exhibit a clear trade-off, whereas our method (red) consistently lies in the upper-right region close to the Teacher (blue).
    }
    \label{fig:diversity_tradeoff}
\end{figure}

\begin{figure*}[t]
  \centering
  \includegraphics[width=\linewidth]{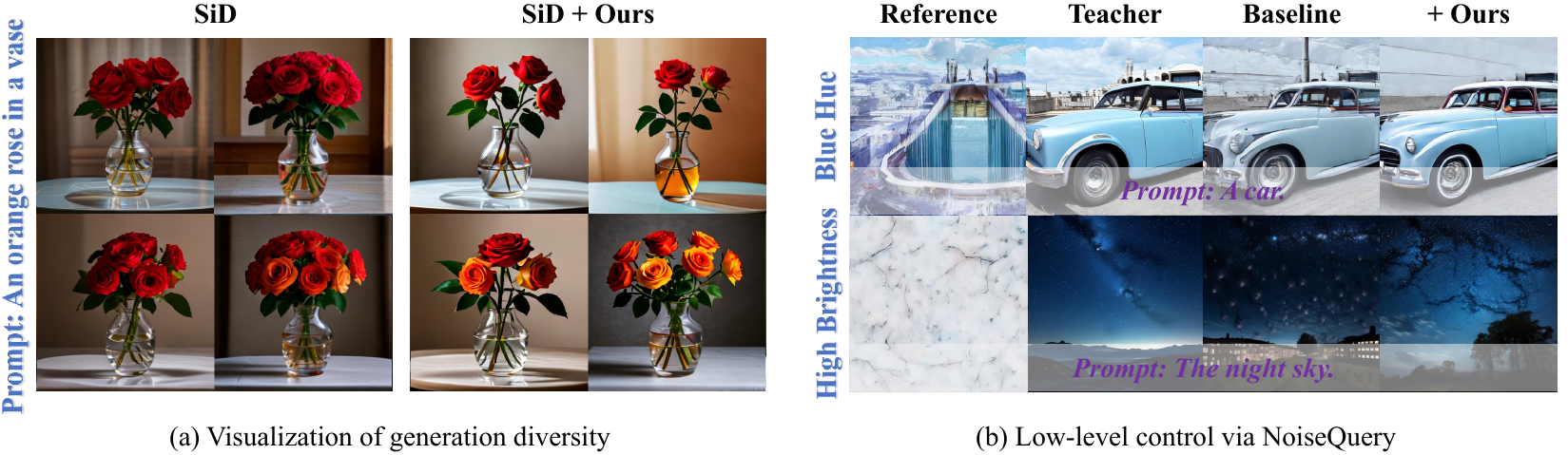}
  \caption{\textbf{Visualization of diversity and low-level control.} (a) Generated images of baseline distilled models (SiD) \cite{zhou2025score} and ours under the same set of initial noises. (b)  Zero-shot control via NoiseQuery \cite{wang2025silent}: retrieving noise for ``Blue Hue" and ``High Brightness" from the teacher. 
  }
  \label{fig:div-low-vis}
  \vspace{-0.3cm}
\end{figure*}

\textbf{Results.}
Fig.~\ref{fig:diversity_tradeoff} illustrates the trade-off landscape. 
Baseline methods often struggle to balance these metrics, either suffering from reduced Vendi Scores or sacrificing CLIP alignment to sustain diversity. In contrast, GAD effectively mitigates this trade-off, consistently residing in the favorable upper-right region near the Teacher. 
Qualitative inspection (Fig.~\ref{fig:div-low-vis} (a)) confirms that GAD produces richer semantic variations, validating that restoring noise sensitivity recovers the teacher's semantic exploration capabilities.

\subsection{Downstream task 3: Transfer of Controllability}
\label{subsec:noise_retrieval}

We further evaluate the functional consistency with the teacher via NoiseQuery \cite{wang2025silent}. 
This method retrieves the optimal initial noise that best matches the target attributes from a pre-computed database.

\textbf{Experimental Protocol.}
We construct a noise-feature database using the teacher model by sampling $10^4$ noise vectors and storing their generated features.
At inference time, we retrieve the optimal noise $\mathbf{z}^*$ based on the teacher feature, and directly apply the retrieved noise to the student model \emph{without any adaptation}.
This establishes a challenging zero-shot transfer setting, where successful control requires the student to share the same noise-to-image geometry as the teacher.
The experiments are conducted on DrawBench \cite{saharia2022photorealistic} dataset. 
We report CLIP Score, HPSv2 \citep{wu2023human}, and PickScore \citep{kirstain2023pick} to measure the alignment and aesthetic quality.

\textbf{Results.}
Quantitative results under semantic retrieval are summarized in Tab.~\ref{tab:noise_query}.
It shows that standard distilled models often fail to benefit from the teacher's optimized noise.
In contrast, applying our method improves performance in most cases.
This confirms that GAD not only restores sensitivity but effectively aligns the student's functional landscape with the teacher to enable zero-shot transfer of test-time enhancement.
Fig.~\ref{fig:div-low-vis}(b) visualizes retrieval for low-level attributes (e.g., blue hue, high brightness).
The leftmost column visualizes the unconditional image generated from $\mathbf{z}^*$, revealing the intrinsic low-level bias encoded in the noise.
Using a minimally informative prompt (e.g., ``\emph{A car}''), our method more faithfully preserves the low-level attributes specified by the retrieved noise.

\begin{table}[t]
\centering
 \caption{\textbf{Quantitative results on zero-shot controlling.} 
Evaluation of control transferability by applying the teacher’s optimal noise retrieved by NoiseQuery to the student during inference. 
 }
\resizebox{\linewidth}{!}{
\begin{tabular}{lccc}
\toprule
\textbf{Model} & \textbf{CLIPScore} $\uparrow$ & \textbf{HPSv2} $\uparrow$  & \textbf{PickScore} $\uparrow$ \\
\midrule
\multicolumn{4}{l}{\textbf{\emph{SD2 (UNet)}}} \\
\rowcolor{lightgray}
Teacher \cite{rombach2022high}      &  31.62 & 0.258  & 21.594 \\
LADD \cite{sauer2024fast}        & 30.79 & 0.223  & 20.839 \\
LADD+Ours    & \textbf{30.87} &  \textbf{0.232}  & \textbf{20.884} \\
\midrule
\multicolumn{4}{l}{\textbf{\emph{PixArt-$\alpha$ (DiT)}}} \\
\rowcolor{lightgray}
Teacher \cite{chenpixart}     & 30.97 & 0.280 & 22.174 \\
TDM \cite{luo2025learning}         & 30.51 & 0.271 & 21.900 \\
TDM+Ours     & \textbf{30.57} & \textbf{0.274} & \textbf{21.980} \\
\midrule
\multicolumn{4}{l}{\textbf{\emph{SANA (Flow-based DiT)}}} \\
\rowcolor{lightgray}
Teacher \cite{xie2025sana}     &  32.33 & 0.289  & 22.523 \\
SiD \cite{zhou2025score}         & \textbf{31.61} &  0.249 & 22.081 \\
SiD+Ours     & 31.59 &  \textbf{0.289} & \textbf{22.190}  \\
\bottomrule
\end{tabular}
}

\label{tab:noise_query}

\end{table}

\subsection{Ablation Study}

\label{subsec:ablation}
\textbf{Strategy Variations.}
To validate the effectiveness of GAD, we compare against two alternative baselines on PixArt-$\alpha$ with 1k COCO prompts.
\textbf{(1) Noise Augmentation:} We train the student with perturbed inputs $\mathbf{z}' = \mathbf{z} + \epsilon \mathbf{v}$ but without the Jacobian matching loss. As shown in Tab.~\ref{tab:ablation}, this setting actually leads to a decrease in Vendi Score. This suggests that without an explicit target response, the student learns to be \textit{invariant} to input noise rather than sensitive to it.
\textbf{(2) Diversity Regularization:} We apply a blind repulsion loss (maximizing $\| \Phi_S(z') - \Phi_S(z) \|$) to force sensitivity. While this yields the highest Vendi Score, it comes at a cost to semantic alignment (CLIP Score drop). This confirms that blindly forcing variance destroys the precise mapping required for high-fidelity generation.
In contrast, GAD acts as a guided sensitivity restoration, where the direction of change is dictated by the teacher's Jacobian, ensuring that diversity is semantically meaningful.
In addition, a detailed sensitivity analysis regarding the perturbation scale $h$ and the weight parameter $\lambda$ can be found in Appendix~\ref{app:ablation}.

\begin{table}[t]
    \centering
    \caption{\textbf{Ablation study on Geometry-Aware Distillation (GAD).} Metrics are reported on PixArt-$\alpha$ with baseline distillation TDM.}
    \label{tab:ablation}
    \resizebox{\linewidth}{!}{
        \begin{tabular}{l|c|ccc}
            \toprule
            \textbf{Method Setting} & \textbf{Detail} & \textbf{Vendi} $\uparrow$ & \textbf{CLIP} $\uparrow$ & \textbf{PickScore} $\uparrow$ \\
            \midrule
            \textbf{Baseline (TDM)} & - &  2.7187 & 33.42 & 22.1240 \\
            \textbf{+ Noise Augmentation} & No constraint & 2.6355 & 33.41 & 22.1924  \\
            \textbf{+ Diversity Reg.} & Blindly Active & \textbf{3.0559} & 31.76 &  21.3556\\
            \midrule
            \textbf{+ Ours} & Geometric Alignment & 2.7967 & \textbf{33.53} & \textbf{22.2741}  \\
            \bottomrule
        \end{tabular}
    }
    \vspace{-0.3cm}
\end{table}

\hhy{\textbf{Supervised Timesteps.}
To understand the temporal sensitivity of geometric alignment, we ablate the timestep intervals where the GAD objective is applied. As shown in Tab.~\ref{tab:ablation_timesteps}, applying GAD exclusively to the middle noise levels ($t \in [200, 600]$) restores a significant portion of generative diversity (Vendi score $2.7140$ vs. baseline $2.5967$). Interestingly, high-noise supervision ($t > 600$) yields the best semantic alignment, while middle-noise supervision is most critical for structural diversity. This temporal disentanglement reveals that GAD does not require dense trajectory alignment to be effective. Notably, the sparse supervision strategy (e.g., mid-noise only) achieves performance highly competitive with the full-step setting, while incidentally offering a significant reduction in the computational overhead associated with the finite-difference approximation.
}

\hhy{\textbf{Perturbation Scale Dynamics.}
We ablate the fixed scale against timestep-adaptive schedules in Tab.~\ref{tab:ablation_h_adaptive}. The results indicate that an increasing linear schedule (i.e., assigning a smaller $h$ at high noise and a larger $h$ at low noise) further enhances both the CLIP score and Vendi diversity. This aligns with the intuition of the reverse generation process: at early stages (high noise), minute perturbations are exponentially amplified by subsequent sampling steps, requiring a finer scale to maintain local linearity. At later stages (low noise), the model focuses on detail refinement and exhibits greater robustness, allowing a larger $h$ to capture local geometric variations more effectively.
}

\begin{table}[t]
\centering
\caption{\hhy{\textbf{Ablation on GAD timesteps.}} The performance of applying GAD to different noise intervals (sparse supervision) on PixArt-$\alpha$ with baseline TDM.}
\label{tab:ablation_timesteps}
\resizebox{\linewidth}{!}{
\begin{tabular}{lcccc}
\toprule
\textbf{GAD Timesteps} & \textbf{CLIP} $\uparrow$ & \textbf{PickScore} $\uparrow$ & \textbf{Vendi} $\uparrow$ \\
\midrule
Baseline (No GAD) & 33.43 & 22.1007 & 2.5967 \\
High noise ($t > 600$) & \textbf{33.59} & 22.2385 & 2.6686 \\
Mid noise ($t \in [200, 600]$) & 33.45 & 22.1274 & 2.7140 \\
Low noise ($t < 200$) & 33.32 & 22.1578 & 2.6680 \\
Random ($20\%$ steps) & 33.10 & 22.1247 & 2.7180 \\
Full steps (Default) & 33.52 & \textbf{22.2704} & \textbf{2.7187} \\
\bottomrule
\end{tabular}
}
\end{table}

\begin{table}[t]
\centering
\caption{\hhy{\textbf{Ablation on perturbation scale dynamics.}} The performance of applying GAD with timestep-adaptive perturbation scale $h$ on PixArt-$\alpha$.}
\label{tab:ablation_h_adaptive}
\small
\resizebox{\linewidth}{!}{
\begin{tabular}{lccc}
\toprule
\textbf{Scale Schedule $h(t)$} & \textbf{CLIP} $\uparrow$ & \textbf{PickScore} $\uparrow$ & \textbf{Vendi} $\uparrow$ \\
\midrule
Linear (decreasing) & 33.32 & 22.2166 & 2.6694 \\
Linear (increasing) & \textbf{33.53} & 22.1427 & \textbf{2.7202} \\
 Fixed (Ours: $h=10^{-2}$) & 33.52 & \textbf{22.2704} & 2.7187 \\
\bottomrule
\end{tabular}
}
\vspace{-0.3cm}
\end{table}

\section{Conclusion}
\label{sec:conclusion}

We identified and formalized \emph{noise sensitivity degradation} in diffusion distillation, showing that standard pointwise objectives preserve perceptual quality but fail to retain the teacher’s local functional geometry, leading to reduced controllability and mode collapse.
To address this issue, we proposed Geometry-Aware Distillation (GAD), which aligns teacher and student models through Jacobian-vector product matching, explicitly preserving their response to input perturbations. Extensive experiments across multiple architectures and distillation paradigms demonstrate that GAD consistently improves functional alignment.
Importantly, restoring noise sensitivity translates into practical gains, including improved layout control, enhanced generative diversity, and more effective transfer of test-time optimization, without compromising image fidelity.

\textbf{Limitations and Future Work.} 
While GAD incurs zero inference cost, our finite-difference approximation introduces additional training overhead (detailed in Appendix~\ref{app:complexity}), as it requires an extra forward pass. However, this remains more efficient than standard second-order methods by reusing cached conditions. Future work could explore more sample-efficient estimators for Jacobian alignment or investigate the theoretical connections between geometric sensitivity and generalization bounds in generative modeling~\cite{miao2023fedseg}. Furthermore, extending GAD to video diffusion distillation~\cite{yang2026towards}, where temporal consistency relies heavily on noise correlations, presents a promising direction.

\section*{Acknowledgments}
\hhy{
This work was partially supported by the National Natural Science Foundation of China under Grant 62372341. YZ was also partially supported by the Hi! PARIS initiative and the ANR/France 2030 program (ANR-23-IACL-0005). Additional support was provided by the MiLM Plus Team at Xiaomi Inc.
}

\section*{Impact Statement}

This work aims to improve accelerated diffusion models by addressing a critical limitation in existing distillation approaches: the loss of sensitivity to initial noise. By preserving this property during acceleration, our method allows fast generative models to retain the diversity, controllability, and stochastic expressiveness of their teacher counterparts, rather than converging to high-quality yet overly homogeneous outputs. 
As with many advances in AI-based generative modeling, this work may have broader societal implications that are common to the deployment of such technologies. From a technical perspective, however, we do not foresee any distinct or additional risks arising from our method beyond those generally associated with current text-to-image generative models.

\bibliography{example_paper}

@inproceedings{nichol2021glide,
  title={Glide: Towards photorealistic image generation and editing with text-guided diffusion models},
  author={Nichol, Alex and Dhariwal, Prafulla and Ramesh, Aditya and Shyam, Pranav and Mishkin, Pamela and McGrew, Bob and Sutskever, Ilya and Chen, Mark},
  booktitle    = {International Conference on Machine Learning},
  volume       = {162},
  pages        = {16784--16804},
  publisher    = {{PMLR}},
  year         = {2022},
}

@inproceedings{lin2014microsoft,
  title={Microsoft coco: Common objects in context},
  author={Lin, Tsung-Yi and Maire, Michael and Belongie, Serge and Hays, James and Perona, Pietro and Ramanan, Deva and Doll{\'a}r, Piotr and Zitnick, C Lawrence},
  booktitle={European Conference on Computer Vision},
  pages={740--755},
  year={2014},
  organization={springer}
}

@article{ho2020denoising,
  title={Denoising diffusion probabilistic models},
  author={Ho, Jonathan and Jain, Ajay and Abbeel, Pieter},
  journal={Advances in Neural Information Processing Systems},
  volume={33},
  pages={6840--6851},
  year={2020}
}

@inproceedings{wang2025silent,
  title={The silent assistant: Noisequery as implicit guidance for goal-driven image generation},
  author={Wang, Ruoyu and Huang, Huayang and Zhu, Ye and Russakovsky, Olga and Wu, Yu},
  booktitle={Proceedings of the IEEE/CVF International Conference on Computer Vision},
  pages={17618--17628},
  year={2025}
}

@inproceedings{zhou2025golden,
  title={Golden noise for diffusion models: A learning framework},
  author={Zhou, Zikai and Shao, Shitong and Bai, Lichen and Zhang, Shufei and Xu, Zhiqiang and Han, Bo and Xie, Zeke},
  booktitle={Proceedings of the IEEE/CVF International Conference on Computer Vision},
  pages={17688--17697},
  year={2025}
}

@article{eyring2024reno,
  title={Reno: Enhancing one-step text-to-image models through reward-based noise optimization},
  author={Eyring, Luca and Karthik, Shyamgopal and Roth, Karsten and Dosovitskiy, Alexey and Akata, Zeynep},
  journal={Advances in Neural Information Processing Systems},
  volume={37},
  pages={125487--125519},
  year={2024}
}

@inproceedings{meng2023distillation,
  title={On distillation of guided diffusion models},
  author={Meng, Chenlin and Rombach, Robin and Gao, Ruiqi and Kingma, Diederik P. and Ermon, Stefano and Ho, Jonathan and Salimans, Tim},
  booktitle={Proceedings of the IEEE/CVF Conference on Computer Vision and Pattern Recognition},
  pages={14297--14306},
  year={2023}
}

@inproceedings{sauer2024adversarial,
  title={Adversarial diffusion distillation},
  author={Sauer, Axel and Lorenz, Dominik and Blattmann, Andreas and Rombach, Robin},
  booktitle={European Conference on Computer Vision},
  pages={87--103},
  year={2024},
  organization={Springer}
}

@inproceedings{xie2023boxdiff,
  title={Boxdiff: Text-to-image synthesis with training-free box-constrained diffusion},
  author={Xie, Jinheng and Li, Yuexiang and Huang, Yawen and Liu, Haozhe and Zhang, Wentian and Zheng, Yefeng and Shou, Mike Zheng},
  booktitle={Proceedings of the IEEE/CVF International Conference on Computer Vision},
  pages={7452--7461},
  year={2023}
}

@inproceedings{bancrystal,
  title={The Crystal Ball Hypothesis in diffusion models: Anticipating object positions from initial noise},
  author={Ban, Yuanhao and Wang, Ruochen and Zhou, Tianyi and Gong, Boqing and Hsieh, Cho-Jui and Cheng, Minhao},
  booktitle={International Conference on Learning Representations},
year={2025}
}

@inproceedings{ciderondiversity,
  title={Diversity-Rewarded CFG Distillation},
  author={Cideron, Geoffrey and Agostinelli, Andrea and Ferret, Johan and Girgin, Sertan and Elie, Romuald and Bachem, Olivier and Perrin, Sarah and Rame, Alexandre},
  booktitle={International Conference on Learning Representations},
year={2025}
}

@inproceedings{song2023consistency,
  title={Consistency Models},
  author={Song, Yang and Dhariwal, Prafulla and Chen, Mark and Sutskever, Ilya},
  booktitle={International Conference on Machine Learning},
  pages={32211--32252},
  year={2023},
  organization={PMLR}
}

@article{gandikota2025distilling,
  title={Distilling diversity and control in diffusion models},
  author={Gandikota, Rohit and Bau, David},
  journal={arXiv preprint arXiv:2503.10637},
  year={2025}
}

@inproceedings{song2020score,
  title={Score-based generative modeling through stochastic differential equations},
  author={Song, Yang and Sohl-Dickstein, Jascha and Kingma, Diederik P and Kumar, Abhishek and Ermon, Stefano and Poole, Ben},
  booktitle    = {International Conference on Learning Representations},
  publisher    = {OpenReview.net},
  year         = {2021},
}

@inproceedings{lipmanflow,
  title={Flow Matching for Generative Modeling},
  author={Lipman, Yaron and Chen, Ricky TQ and Ben-Hamu, Heli and Nickel, Maximilian and Le, Matthew},
  booktitle={International Conference on Learning Representations},
year={2023}
}

@article{lin2024sdxl,
  title={Sdxl-lightning: Progressive adversarial diffusion distillation},
  author={Lin, Shanchuan and Wang, Anran and Yang, Xiao},
  journal={arXiv preprint arXiv:2402.13929},
  year={2024}
}

@inproceedings{sauer2024fast,
  title={Fast high-resolution image synthesis with latent adversarial diffusion distillation},
  author={Sauer, Axel and Boesel, Frederic and Dockhorn, Tim and Blattmann, Andreas and Esser, Patrick and Rombach, Robin},
  booktitle={SIGGRAPH Asia},
  pages={1--11},
  year={2024}
}

@inproceedings{yin2024one,
  title={One-step diffusion with distribution matching distillation},
  author={Yin, Tianwei and Gharbi, Micha{\"e}l and Zhang, Richard and Shechtman, Eli and Durand, Fredo and Freeman, William T and Park, Taesung},
  booktitle={Proceedings of the IEEE/CVF Conference on Computer Vision and Pattern Recognition},
  pages={6613--6623},
  year={2024}
}

@article{yin2024improved,
  title={Improved distribution matching distillation for fast image synthesis},
  author={Yin, Tianwei and Gharbi, Micha{\"e}l and Park, Taesung and Zhang, Richard and Shechtman, Eli and Durand, Fredo and Freeman, Bill},
  journal={Advances in Neural Information Processing Systems},
  volume={37},
  pages={47455--47487},
  year={2024}
}

@inproceedings{songimproved,
  title={Improved Techniques for Training Consistency Models},
  author={Song, Yang and Dhariwal, Prafulla},
  booktitle={International Conference on Learning Representations},
year={2024}
}

@inproceedings{srinivas2018knowledge,
  title={Knowledge transfer with jacobian matching},
  author={Srinivas, Suraj and Fleuret, Fran{\c{c}}ois},
  booktitle={International Conference on Machine Learning},
  pages={4723--4731},
  year={2018},
  organization={PMLR}
}

@inproceedings{humayunsecrets,
  title={What Secrets Do Your Manifolds Hold? Understanding the Local Geometry of Generative Models},
  author={Humayun, Ahmed Imtiaz and Amara, Ibtihel and Vasconcelos, Cristina Nader and Ramachandran, Deepak and Schumann, Candice and He, Junfeng and Heller, Katherine A and Farnadi, Golnoosh and Rostamzadeh, Negar and Havaei, Mohammad},
  booktitle={International Conference on Learning Representations},
year={2025}
}

@inproceedings{umboost,
  title={Boost-and-Skip: A Simple Guidance-Free Diffusion for Minority Generation},
  author={Um, Soobin and Kim, Beomsu and Ye, Jong Chul},
  booktitle={International Conference on Machine Learning},
year={2025}
}

@inproceedings{um2025minority,
  title={Minority-Focused Text-to-Image Generation via Prompt Optimization},
  author={Um, Soobin and Ye, Jong Chul},
  booktitle={Proceedings of the IEEE/CVF Conference on Computer Vision and Pattern Recognition},
  pages={20926--20936},
  year={2025}
}

@inproceedings{sadatcads,
  title={CADS: Unleashing the Diversity of Diffusion Models through Condition-Annealed Sampling},
  author={Sadat, Seyedmorteza and Buhmann, Jakob and Bradley, Derek and Hilliges, Otmar and Weber, Romann M},
  booktitle={International Conference on Learning Representations},
year={2024}
}

@inproceedings{li2024enhancing,
  title={Enhancing compositional text-to-image generation with reliable random seeds},
  author={Li, Shuangqi and Le, Hieu and Xu, Jingyi and Salzmann, Mathieu},
  booktitle={International Conference on Learning Representations},
  year={2025}
}

@inproceedings{luo2025learning,
  title={Learning Few-Step Diffusion Models by Trajectory Distribution Matching},
  author={Luo, Yihong and Hu, Tianyang and Sun, Jiacheng and Cai, Yujun and Tang, Jing},
  booktitle={Proceedings of the IEEE/CVF International Conference on Computer Vision},
pages     = {17719-17728},
  year={2025}
}

@article{pearlmutter1994fast,
  title={Fast exact multiplication by the Hessian},
  author={Pearlmutter, Barak A},
  journal={Neural computation},
  volume={6},
  number={1},
  pages={147--160},
  year={1994},
  publisher={MIT Press}
}

@article{baydin2018automatic,
  title={Automatic differentiation in machine learning: a survey},
  author={Baydin, Atilim Gunes and Pearlmutter, Barak A and Radul, Alexey Andreyevich and Siskind, Jeffrey Mark},
  journal={JMLR},
  volume={18},
  number={153},
  pages={1--43},
  year={2018}
}

@inproceedings{finlay2020train,
  title={How to train your neural ode: the world of jacobian and kinetic regularization},
  author={Finlay, Chris and Jacobsen, J{\"o}rn-Henrik and Nurbekyan, Levon and Oberman, Adam},
  booktitle={International Conference on Machine Learning},
  pages={3154--3164},
  year={2020},
  organization={PMLR}
}

@inproceedings{caesar2018coco,
  title={Coco-stuff: Thing and stuff classes in context},
  author={Caesar, Holger and Uijlings, Jasper and Ferrari, Vittorio},
  booktitle={Proceedings of the IEEE/CVF Conference on Computer Vision and Pattern Recognition},
  pages={1209--1218},
  year={2018}
}

@article{bochkovskiy2020yolov4,
  title={Yolov4: Optimal speed and accuracy of object detection},
  author={Bochkovskiy, Alexey and Wang, Chien-Yao and Liao, Hong-Yuan Mark},
  journal={arXiv preprint arXiv:2004.10934},
  year={2020}
}

@inproceedings{li2021image,
  title={Image synthesis from layout with locality-aware mask adaption},
  author={Li, Zejian and Wu, Jingyu and Koh, Immanuel and Tang, Yongchuan and Sun, Lingyun},
  booktitle={Proceedings of the IEEE/CVF International Conference on Computer Vision},
  pages={13819--13828},
  year={2021}
}

@inproceedings{radford2021learning,
  title={Learning transferable visual models from natural language supervision},
  author={Radford, Alec and Kim, Jong Wook and Hallacy, Chris and Ramesh, Aditya and Goh, Gabriel and Agarwal, Sandhini and Sastry, Girish and Askell, Amanda and Mishkin, Pamela and Clark, Jack and others},
  booktitle={International Conference on Machine Learning},
  pages={8748--8763},
  year={2021},
  organization={PmLR}
}

@inproceedings{rombach2022high,
  title={High-resolution image synthesis with latent diffusion models},
  author={Rombach, Robin and Blattmann, Andreas and Lorenz, Dominik and Esser, Patrick and Ommer, Bj{\"o}rn},
  booktitle={Proceedings of the IEEE/CVF Conference on Computer Vision and Pattern Recognition},
  pages={10684--10695},
  year={2022}
}

@inproceedings{nguyen2024swiftbrush,
  title={Swiftbrush: One-step text-to-image diffusion model with variational score distillation},
  author={Nguyen, Thuan Hoang and Tran, Anh},
  booktitle={Proceedings of the IEEE/CVF Conference on Computer Vision and Pattern Recognition},
  pages={7807--7816},
  year={2024}
}

@inproceedings{dao2024swiftbrush,
  title={Swiftbrush v2: Make your one-step diffusion model better than its teacher},
  author={Dao, Trung and Nguyen, Thuan Hoang and Le, Thanh and Vu, Duc and Nguyen, Khoi and Pham, Cuong and Tran, Anh},
  booktitle={European Conference on Computer Vision},
  pages={176--192},
  year={2024},
  organization={Springer}
}

@article{luo2023latent,
  title={Latent consistency models: Synthesizing high-resolution images with few-step inference},
  author={Luo, Simian and Tan, Yiqin and Huang, Longbo and Li, Jian and Zhao, Hang},
  journal={arXiv preprint arXiv:2310.04378},
  year={2023}
}

@article{zheng2024trajectory,
  title={Trajectory Consistency Distillation: Improved Latent Consistency Distillation by Semi-Linear Consistency Function with Trajectory Mapping},
  author={Zheng, Jianbin and Hu, Minghui and Fan, Zhongyi and Wang, Chaoyue and Ding, Changxing and Tao, Dacheng and Cham, Tat-Jen},
  journal={arXiv preprint arXiv:2402.19159},
  year={2024}
}

@inproceedings{changpinyo2021conceptual,
  title={Conceptual 12m: Pushing web-scale image-text pre-training to recognize long-tail visual concepts},
  author={Changpinyo, Soravit and Sharma, Piyush and Ding, Nan and Soricut, Radu},
  booktitle={Proceedings of the IEEE/CVF Conference on Computer Vision and Pattern Recognition},
  pages={3558--3568},
  year={2021}
}

@article{dan2023vendi,
  title={The Vendi Score: A Diversity Evaluation Metric for Machine Learning},
  author={Dan Friedman, Dan and Dieng, Adji Bousso},
  journal={Transactions on Machine Learning Research},
issn={2835-8856},
  year={2023}
}

@inproceedings{chenpixart,
  title={PixArt-$\alpha$: Fast Training of Diffusion Transformer for Photorealistic Text-to-Image Synthesis},
  author={Chen, Junsong and Jincheng, YU and Chongjian, GE and Yao, Lewei and Xie, Enze and Wang, Zhongdao and Kwok, James and Luo, Ping and Lu, Huchuan and Li, Zhenguo},
  booktitle={International Conference on Learning Representations},
year={2024}
}

@inproceedings{xie2025sana,
  title={SANA: Efficient high-resolution text-to-image synthesis with linear diffusion transformers},
  author={Xie, Enze and Chen, Junsong and Chen, Junyu and Cai, Han and Tang, Haotian and Lin, Yujun and Zhang, Zhekai and Li, Muyang and Zhu, Ligeng and Lu, Yao and others},
  booktitle={International Conference on Learning Representations},
  year={2025}
}

@article{zhou2025score,
  title={Score Distillation of Flow Matching Models},
  author={Zhou, Mingyuan and Gu, Yi and Zheng, Huangjie and Song, Liangchen and He, Guande and Zhang, Yizhe and Hu, Wenze and Yang, Yinfei},
  journal={arXiv preprint arXiv:2509.25127},
  year={2025}
}

@inproceedings{chadebec2025flash,
  title={Flash diffusion: Accelerating any conditional diffusion model for few steps image generation},
  author={Chadebec, Clement and Tasar, Onur and Benaroche, Eyal and Aubin, Benjamin},
  booktitle={Proceedings of the AAAI Conference on Artificial Intelligence},
  volume={39},
  number={15},
  pages={15686--15695},
  year={2025}
}

@inproceedings{luoyou,
  title={You Only Sample Once: Taming One-Step Text-to-Image Synthesis by Self-Cooperative Diffusion GANs},
  author={Luo, Yihong and Chen, Xiaolong and Qu, Xinghua and Hu, Tianyang and Tang, Jing},
  booktitle={International Conference on Learning Representations},
year={2025}
}

@article{saharia2022photorealistic,
  title={Photorealistic text-to-image diffusion models with deep language understanding},
  author={Saharia, Chitwan and Chan, William and Saxena, Saurabh and Li, Lala and Whang, Jay and Denton, Emily L and Ghasemipour, Kamyar and Gontijo Lopes, Raphael and Karagol Ayan, Burcu and Salimans, Tim and others},
  journal={Advances in Neural Information Processing Systems},
  volume={35},
  pages={36479--36494},
  year={2022}
}

@article{wu2023human,
  title={Human preference score v2: A solid benchmark for evaluating human preferences of text-to-image synthesis},
  author={Wu, Xiaoshi and Hao, Yiming and Sun, Keqiang and Chen, Yixiong and Zhu, Feng and Zhao, Rui and Li, Hongsheng},
  journal={arXiv preprint arXiv:2306.09341},
  year={2023}
}

@article{kirstain2023pick,
  title={Pick-a-pic: An open dataset of user preferences for text-to-image generation},
  author={Kirstain, Yuval and Polyak, Adam and Singer, Uriel and Matiana, Shahbuland and Penna, Joe and Levy, Omer},
  journal={Advances in Neural Information Processing Systems},
  volume={36},
  pages={36652--36663},
  year={2023}
}

@article{li2022efficientformer,
  title={Efficientformer: Vision transformers at mobilenet speed},
  author={Li, Yanyu and Yuan, Geng and Wen, Yang and Hu, Ju and Evangelidis, Georgios and Tulyakov, Sergey and Wang, Yanzhi and Ren, Jian},
  journal={Advances in Neural Information Processing Systems},
  volume={35},
  pages={12934--12949},
  year={2022}
}

@inproceedings{liuflow,
  title={Flow Straight and Fast: Learning to Generate and Transfer Data with Rectified Flow},
  author={Liu, Xingchao and Gong, Chengyue and others},
  booktitle={International Conference on Learning Representations},
year={2023}
}

@article{hutchinson1989stochastic,
  title={A stochastic estimator of the trace of the influence matrix for Laplacian smoothing splines},
  author={Hutchinson, Michael F},
  journal={Communications in Statistics-Simulation and Computation},
  volume={18},
  number={3},
  pages={1059--1076},
  year={1989},
  publisher={Taylor \& Francis}
}

@article{czarnecki2017sobolev,
  title={Sobolev training for neural networks},
  author={Czarnecki, Wojciech M and Osindero, Simon and Jaderberg, Max and Swirszcz, Grzegorz and Pascanu, Razvan},
  journal={Advances in Neural Information Processing Systems},
  volume={30},
  year={2017}
}

@inproceedings{kimconsistency,
  title={Consistency Trajectory Models: Learning Probability Flow ODE Trajectory of Diffusion},
  author={Kim, Dongjun and Lai, Chieh-Hsin and Liao, Wei-Hsiang and Murata, Naoki and Takida, Yuhta and Uesaka, Toshimitsu and He, Yutong and Mitsufuji, Yuki and Ermon, Stefano},
  booktitle={International Conference on Learning Representations},
year={2024}
}

@inproceedings{lu2025adversarial,
  title={Adversarial distribution matching for diffusion distillation towards efficient image and video synthesis},
  author={Lu, Yanzuo and Ren, Yuxi and Xia, Xin and Lin, Shanchuan and Wang, Xing and Xiao, Xuefeng and Ma, Andy J and Xie, Xiaohua and Lai, Jian-Huang},
  booktitle={Proceedings of the IEEE/CVF International Conference on Computer Vision},
  pages={16818--16829},
  year={2025}
}

@inproceedings{chen2025nitrofusion,
  title={Nitrofusion: High-fidelity single-step diffusion through dynamic adversarial training},
  author={Chen, Dar-Yen and Bandyopadhyay, Hmrishav and Zou, Kai and Song, Yi-Zhe},
  booktitle={Proceedings of the IEEE/CVF Conference on Computer Vision and Pattern Recognition},
  pages={7654--7663},
  year={2025}
}

@inproceedings{nguyen2025supercharged,
  title={Supercharged One-step Text-to-Image Diffusion Models with Negative Prompts},
  author={Nguyen, Viet and Nguyen, Anh and Dao, Trung and Nguyen, Khoi and Pham, Cuong and Tran, Toan and Tran, Anh},
  booktitle={Proceedings of the IEEE/CVF International Conference on Computer Vision},
  pages={18004--18013},
  year={2025}
}

@article{starodubcev2024invertible,
  title={Invertible consistency distillation for text-guided image editing in around 7 steps},
  author={Starodubcev, Nikita and Khoroshikh, Mikhail and Babenko, Artem and Baranchuk, Dmitry},
  journal={Advances in Neural Information Processing Systems},
  volume={37},
  pages={12496--12527},
  year={2024}
}

@inproceedings{
kirchhof2025shielded,
title={Shielded Diffusion: Generating Novel and Diverse Images using Sparse Repellency},
author={Michael Kirchhof and James Thornton and Louis B{\'e}thune and Pierre Ablin and Eugene Ndiaye and marco cuturi},
booktitle={International Conference on Machine Learning},
pages = 	 {30911--30942},
volume = 	 {267},
year={2025},
}

@article{zhou2025few,
  title={Few-step diffusion via score identity distillation},
  author={Zhou, Mingyuan and Gu, Yi and Wang, Zhendong},
  journal={arXiv preprint arXiv:2505.12674},
  year={2025}
}

@inproceedings{zhouguided,
  title={Guided Score identity Distillation for Data-Free One-Step Text-to-Image Generation},
  author={Zhou, Mingyuan and Wang, Zhendong and Zheng, Huangjie and Huang, Hai},
  booktitle={International Conference on Learning Representations},
year={2025}
}

@inproceedings{park2019relational,
  title={Relational knowledge distillation},
  author={Park, Wonpyo and Kim, Dongju and Lu, Yan and Cho, Minsu},
  booktitle={IEEE/CVF Conference on Computer Vision and Pattern Recognition},
  pages={3967--3976},
  year={2019}
}

@inproceedings{tung2019similarity,
  title={Similarity-preserving knowledge distillation},
  author={Tung, Frederick and Mori, Greg},
  booktitle={IEEE/CVF International Conference on Computer Vision},
  pages={1365--1374},
  year={2019}
}

@inproceedings{miao2023fedseg,
  title={Fedseg: Class-heterogeneous federated learning for semantic segmentation},
  author={Miao, Jiaxu and Yang, Zongxin and Fan, Leilei and Yang, Yi},
  booktitle={Proceedings of the IEEE/CVF conference on computer vision and pattern recognition},
  pages={8042--8052},
  year={2023}
}

@inproceedings{
yang2026towards,
title={Towards One-step Causal Video Generation via Adversarial Self-Distillation},
author={Yongqi Yang and Huayang Huang and Xu Peng and Xiaobin Hu and Donghao Luo and Jiangning Zhang and Chengjie Wang and Yu Wu},
booktitle={International Conference on Learning Representations},
year={2026},
}
\bibliographystyle{icml2026}

\newpage
\appendix
\onecolumn


\section*{Appendices}
\hhy{The appendix is structured as follows: First, Sec.~\ref{app:ablation} provides comprehensive ablation studies and parameter settings, analyzing the method's sensitivity to key hyperparameters like the perturbation scale and weighting parameter. Next, Sec.~\ref{app:complexity} evaluates the computational complexity and training efficiency of our proposed GAD method across different backbones. Sec.~\ref{app:baselines} details the representative baseline distillation methods used for comparison, while Sec.~\ref{app:impl_details} provides detailed implementation specifications to ensure reproducibility. Furthermore, Sec.~\ref{sec:appendix_fid} presents additional quantitative evaluations focused on zero-shot FID scores. Finally, Sec.~\ref{sec:swiss_roll} offers a 2D toy example using the Swiss Roll dataset to intuitively visualize the geometric gap, and Sec.~\ref{app:visualize} includes more qualitative visualization results demonstrating improvements in diversity and noise-based layout control.
}

\section{Ablation Study and Parameter Settings}
\label{app:ablation}
\textbf{Experimental Setup.} We evaluate the sensitivity of GAD to key hyperparameters using PixArt-$\alpha$ (DiT). All ablation experiments are conducted over 2k training steps. To monitor performance dynamically, metrics are computed online using 50 randomly sampled MS-COCO prompts, with 4 images generated per prompt to calculate intra-prompt LPIPS.

\begin{figure*}[h]
  \centering
  \includegraphics[width=\linewidth]{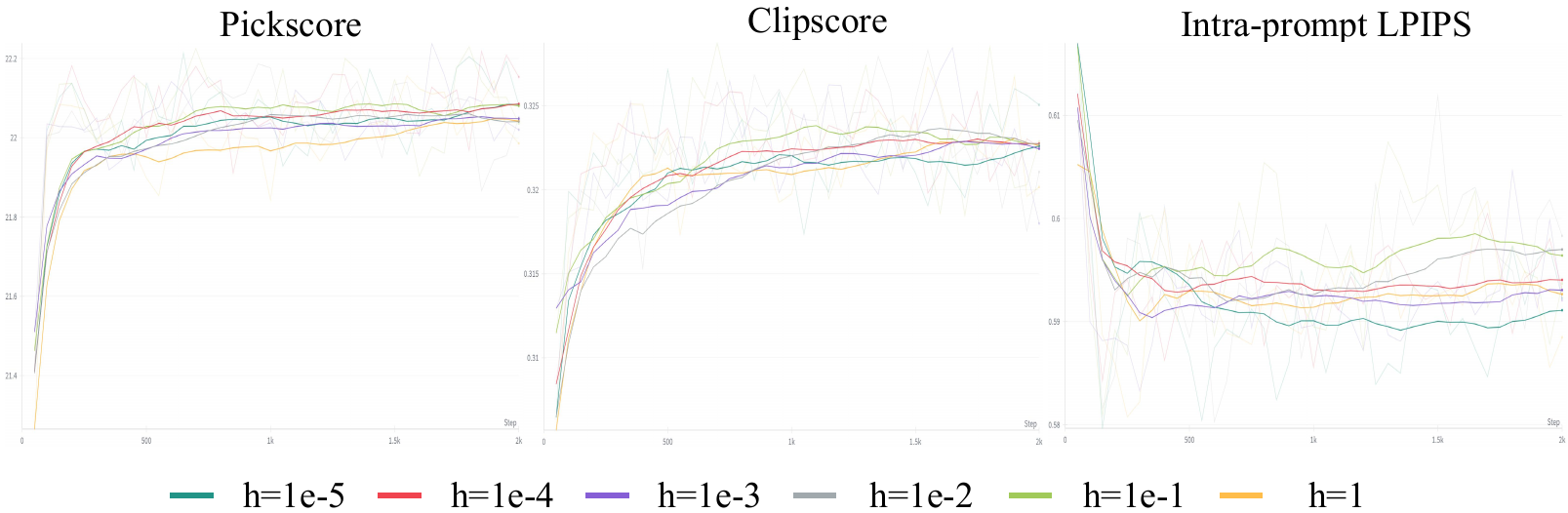}
  \caption{\textbf{Ablation study on the perturbation scale $h$}. Training curves on PixArt-$\alpha$ for Pickscore (left), CLIP Score (middle), and Intra-prompt LPIPS (right). The results indicate that very small $h$ values fail to restore diversity (LPIPS), while $h=10^{-2}$ (grey) achieves an optimal equilibrium between structural sensitivity and generation quality.}
  \label{fig:ablation_h}
\end{figure*}

\textbf{Sensitivity to Perturbation Scale $h$.}
We investigate the impact of the finite difference interval $h \in \{10^{-5}, \dots, 1\}$. As illustrated in Fig.~\ref{fig:ablation_h}, the choice of $h$ involves a critical trade-off between capturing structural trends and maintaining local linearity. When $h$ is too small (e.g., $10^{-5}$), the regularization signal becomes numerically insignificant, resulting in a collapse toward repetitive modes as indicated by the lowest intra-prompt LPIPS (green line). Conversely, an excessively large scale (e.g., $h=1$) violates the local tangent space assumption, degrading the learned manifold and leading to sub-optimal fidelity. We select $h=10^{-2}$ (grey line) as the default, as it provides the most stable convergence and an optimal equilibrium between curvature preservation and generation quality.

\begin{figure*}[h]
  \centering
  \includegraphics[width=\linewidth]{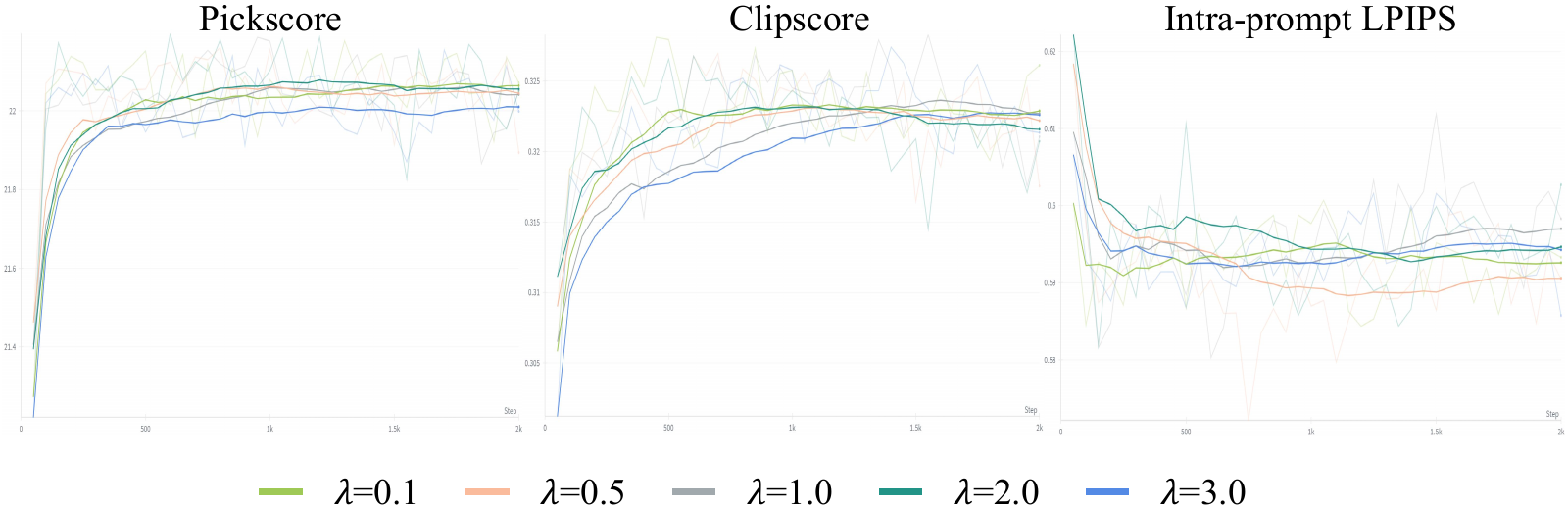}
  \caption{\textbf{Ablation study on the weighting parameter $\lambda$}. Training curves on PixArt-$\alpha$ for Pickscore (left), CLIP Score (middle), and Intra-prompt LPIPS (right). Metrics are computed online during training on a subset of 50 COCO prompts. The results highlight a trade-off: high $\lambda$ values slightly compromise fidelity scores, while low $\lambda$ values lead to diminished LPIPS (diversity), with $\lambda=1.0$ yielding the most balanced performance.
  }
  \label{fig:ablation_lambda}
\end{figure*}

\textbf{Sensitivity to Weighted Parameter $\lambda$.}
We further analyze the balance between the base distillation loss and geometric regularization by varying $\lambda \in \{0.1, \dots, 3.0\}$. GAD demonstrates strong robustness across this range (Fig.~\ref{fig:ablation_lambda}). Specifically, an excessively large weight (e.g., $\lambda=3.0$) slightly interferes with the primary pointwise objective, causing a marginal decline in PickScore and CLIP Score. In contrast, a small weight (e.g., $\lambda=0.1$) provides insufficient regularization to fully restore the model's noise sensitivity, leading to diminished LPIPS. The setting $\lambda=1.0$ (grey line) serves as the optimal balance, effectively reconciling the teacher’s local geometry with high semantic alignment.

\section{Complexity and Training Efficiency}
\label{app:complexity}

To evaluate the computational footprint of GAD, we analyze the training overhead across three representative distillation settings. As GAD employs a finite-difference approximation, it primarily introduces one additional forward pass per training step. Importantly, the gradient is only back-propagated through the student model, and the teacher/auxiliary estimators remain in constant memory during the GAD loss computation.

In Table \ref{tab:overhead}, we report the wall-clock time per iteration and peak GPU memory usage during training for the baseline methods versus their GAD-enhanced versions. Benchmarks for LADD (SD2.1) were performed on NVIDIA A800 (80GB) GPUs, while TDM (PixArt-$\alpha$) and SiD (SANA) were benchmarked on NVIDIA RTX 4090 (48GB) GPUs.

\begin{table}[h]
\centering
\caption{\textbf{Detailed computational overhead.} We compare the baseline training cost with the GAD-integrated version. Time refers to the wall-clock duration per training iteration. Memory refers to the peak VRAM usage per GPU during training.}
\label{tab:overhead}
\begin{tabular}{lcccccc}
\toprule
\multirow{2}{*}{\textbf{Method (Model)}} & \multirow{2}{*}{\textbf{Resolution}} & \multicolumn{2}{c}{\textbf{Training Time (s / iter)}} & & \multicolumn{2}{c}{\textbf{Peak Memory (GB)}} \\
\cmidrule{3-4} \cmidrule{6-7}
& & Baseline & + GAD ($\%$) & & Baseline & + GAD ($\%$) \\ 
\midrule
LADD (SD2.1) & $512^2$ & 20.89 & 33.77 (+61.66\%) & & 47.45 & 69.67 (+46.82\%) \\
TDM (PixArt-$\alpha$) & $512^2$ & 4.38 & 5.97 (+36.30\%) & & 31.44 & 32.61 (+3.72\%) \\
SiD (SANA) & $512^2$ & 48.34 & 80.49 (+66.51\%) & & 34.08 & 40.31 (+18.28\%) \\
\bottomrule
\end{tabular}
\end{table}

The empirical results demonstrate that GAD introduces a marginal overhead. The relative increase in training time is significantly lower than a factor of 2$\times$ (which a naive second-order method would require), because the perturbation $\Delta \epsilon$ reuses the cached conditions and latents from the primary forward pass. Moreover, since GAD is only active during training, it incurs \textbf{zero additional cost} during inference.

\section{Details of Baseline Distillation Methods}
\label{app:baselines}

To comprehensively evaluate the proposed GAD, we compare against a diverse set of representative diffusion distillation paradigms spanning three major backbones: Stable Diffusion v2.1 (U-Net), PixArt-$\alpha$ (DiT), and SANA (Flow-based DiT). 
These baselines cover adversarial distillation, consistency modeling, distribution matching, and trajectory-level alignment, reflecting the dominant directions in recent diffusion acceleration research.

\subsection{Distillation for SD2.1 (U-Net)}
\begin{itemize}

    \item \textbf{SD-Turbo \cite{sauer2024adversarial}:} SD-Turbo is trained using \emph{Adversarial Diffusion Distillation (ADD)}, which combines two primary objectives: (i) an adversarial loss that utilizes a discriminator to ensure the generated samples align with the manifold of real images, and (ii) a distillation loss that leverages a frozen teacher diffusion model to maintain the original model's extensive knowledge. By integrating these losses, SD-Turbo achieves high-quality one-step generation without the need for classifier-free guidance during inference. In our experiments, we use the publicly available checkpoint from Hugging Face\footnote{\url{https://huggingface.co/stabilityai/sd-turbo}}.

    \item \textbf{SwiftBrush \& SwiftBrushv2 \cite{nguyen2024swiftbrush, dao2024swiftbrush}:} 
    SwiftBrush formulates distillation as \emph{variational score distillation}, transferring a pretrained text-to-image prior into a lightweight student by matching predicted noise distributions under a variational objective.
    SwiftBrushv2 further refines the objective with improved initialization and stability techniques, emphasizing faithful reconstruction of teacher outputs in few-step regimes. For our evaluation, we employ the publicly released checkpoints of SwiftBrush\footnote{\url{https://huggingface.co/thuanz123/swiftbrush}} and SwiftBrushv2\footnote{\url{https://drive.google.com/drive/folders/1eUVwTrkOVWT2gCJ4TiWlZmCV2sODuvQD}}.

    \item \textbf{Trajectory Consistency Distillation (TCD) \cite{zheng2024trajectory}:} 
    TCD extends consistency models by explicitly enforcing \emph{multi-step trajectory consistency} between intermediate denoising states.
    The student is trained such that predictions at different timesteps remain coherent under teacher-guided transitions, improving sample quality for small numbers of sampling steps while implicitly regularizing the temporal structure of the diffusion process. In our experiments, we use the publicly available TCD-SD21-base LoRA checkpoint\footnote{\url{https://huggingface.co/h1t/TCD-SD21-base-LoRA}}.

    \item \textbf{LADD \cite{sauer2024fast}:} 
    Latent Adversarial Diffusion Distillation introduces adversarial supervision in the \emph{latent space} rather than pixel space in adversarial diffusion distillation, using a strong teacher to provide high-level semantic feedback.
    By aligning latent distributions through a discriminator, LADD prioritizes semantic fidelity and perceptual realism. In our study, LADD serves as the foundational framework for implementing GAD on SD2. To ensure strict variable control and a fair comparison, we retrained the models both with and without GAD regularization, utilizing the publicly available Nitro-1 codebase\footnote{\url{https://github.com/AMD-AGI/Nitro-1}}.
\end{itemize}

\subsection{Distillation for PixArt-$\alpha$ (DiT)}
\begin{itemize}
    \item \textbf{Latent Consistency Models (LCM) \cite{luo2023latent}:} 
    LCM trains the student to map any intermediate latent state along the probability flow ODE directly to the clean data point.
    This formulation collapses the entire denoising trajectory into a single consistency constraint, allowing flexible few-step sampling while sacrificing explicit modeling of trajectory geometry. In our evaluation, we utilize the publicly available PixArt-LCM-XL-2 checkpoint\footnote{\url{https://huggingface.co/PixArt-alpha/PixArt-LCM-XL-2-1024-MS}}.
    
    \item \textbf{Distribution Matching Distillation (DMD) \cite{yin2024one}:} 
    DMD frames distillation as minimizing the KL divergence between student-generated samples and the real data distribution.
    It employs two score estimators-one from a pretrained diffusion model and one from a learned critic-to approximate the density ratio, enabling few-step generation without requiring paired teacher trajectories. In our experiments, we use the publicly available PixArt-Alpha DMD checkpoint\footnote{\url{https://huggingface.co/PixArt-alpha/PixArt-Alpha-DMD-XL-2-512x512}}.

    \item \textbf{FLASH \cite{chadebec2025flash}:} Flash Diffusion (FLASH) is a versatile distillation method designed to maintain high image quality under drastic step reduction. The student is trained to predict, in a single step, a multi-step denoised output from a teacher model. This process is supervised by a combination of: (i) a distillation loss between student and teacher predictions, (ii) an adversarial objective to drive the student distribution toward the real image manifold, and (iii) a distribution matching (DMD) loss to prevent drifting from the teacher's learned distribution. For our experiments, we utilize the publicly released FLASH-PixArt checkpoint\footnote{\url{https://huggingface.co/jasperai/flash-pixart}}.

    \item \textbf{YOSO \cite{luoyou}:} YOSO (\emph{You Only Sample Once}) is a one-step generation framework that integrates diffusion processes with GANs while addressing the training instability and mode collapse common in adversarial distillation. The core mechanism is a self-cooperative learning strategy that smoothes the adversarial divergence by using the denoising generator itself: it treats one-step generation from less corrupted samples as the ground truth to supervise the generation from more corrupted samples. For text-to-image synthesis, YOSO incorporates several specialized techniques, including latent perceptual loss, Informative Prior Initialization (IPI), and a quick adaptation stage to refine the noise scheduler. In our experiments, we utilize the publicly available YOSO-PixArt-512 checkpoint\footnote{\url{https://huggingface.co/Luo-Yihong/yoso_pixart512}}.

    \item \textbf{Trajectory Distribution Matching (TDM) \cite{luo2025learning}:} 
    TDM aligns the \emph{entire sampling trajectory distribution} of the student with that of the teacher by matching probability flow ODE paths.
    By enforcing distribution consistency across timesteps, TDM aims to preserve global diffusion behavior beyond marginal sample quality. In our study, TDM serves as the foundational distillation framework for our experiments on PixArt-$\alpha$. To ensure a fair comparison and strict variable control, we utilized the official TDM codebase\footnote{\url{https://github.com/Luo-Yihong/TDM}} to retrain the models both with and without GAD regularization. 
\end{itemize}

\subsection{Distillation for SANA (Flow-based DiT)}
\begin{itemize}
    \item \textbf{Score Identity Distillation (SiD) \cite{zhou2025score}:} SiD provides a unified distillation framework for both Gaussian diffusion and flow matching by leveraging a simplified derivation based on Bayes' rule and conditional expectations. It demonstrates that score distillation can be applied broadly to text-to-image flow-matching models without requiring teacher fine-tuning or architectural changes. In our study, SiD serves as the foundational distillation framework for implementing GAD on SANA. To ensure strict variable control and a fair comparison, we utilized the official SiD codebase\footnote{\url{https://github.com/apple/ml-sid-dit/}} to retrain the distillation models both with and without GAD regularization.
\end{itemize}

\section{Detailed Implementation Specifications}
\label{app:impl_details}

This section provides additional technical details regarding the network architectures, hyperparameter settings, and the training procedure to ensure reproducibility.
Across all distillation settings, we employ the AdamW optimizer with $\beta_1=0.9$ and $\beta_2=0.999$. The balancing weight $\lambda$ is set to $1.0$ across all models. Specific parameters for each backbone are summarized in Table \ref{tab:hyperparams}.

\begin{table}[h]
\centering
\caption{\textbf{Key training hyperparameters.} Detailed configurations for GAD integration across different backbones.}
\label{tab:hyperparams}
\begin{tabular}{lccc}
\toprule
\textbf{Configuration} & \textbf{SD2.1 (U-Net)} & \textbf{PixArt-$\alpha$ (DiT)} & \textbf{SANA (Flow-based DiT)} \\
\midrule
Learning Rate & $1 \times 10^{-6}$ & $4 \times 10^{-6}$ & $5 \times 10^{-6}$ \\
Total Iterations & 100K & 2K & 5K \\
Batch Size (per GPU) & 64 & 64 & 8 \\
Precision & BF16 & FP16 & BF16 \\
Perturbation Scale $h$ & 0.0001 & 0.01 & 0.01 \\
GAD Weight $\lambda$ & 1.0 & 1.0 & 1.0 \\
\bottomrule
\end{tabular}
\end{table}

\hhy{
\section{Additional Quantitative Evaluation: FID Scores}
\label{sec:appendix_fid}
}
In Sect.~\ref{subsec:general_quality} of the main text, we evaluate the general generation quality using PickScore and CLIP Score to assess image fidelity and text-image alignment. To further address how the models fit the overall data distribution, we report the zero-shot Fréchet Inception Distance (FID) on the MS-COCO 30K dataset in Tab.~\ref{tab:fid_scores}.

The results demonstrate that GAD consistently enhances distributional alignment with the teacher model, evidenced by the lower ``FID vs.\ Teacher'' scores across all architectures. Notably, in configurations where the baseline distillation method (e.g., TDM on PixArt-$\alpha$) happens to achieve a lower FID than the teacher itself, adding GAD leads to a marginal increase in the ``FID vs.\ COCO-30k'' metric. This phenomenon aligns with our core objective: GAD prioritizes capturing the teacher's underlying geometric distribution rather than merely optimizing for external dataset metrics. Consequently, the student's output distribution shifts to be more ``teacher-like,'' pulling the FID closer to the teacher's original performance. This further validates that GAD effectively restores the original generative dynamics and distribution of the teacher model.

\begin{table}[ht]
\centering
\caption{\textbf{Quantitative comparison of zero-shot FID scores on the MS-COCO 30K dataset.} Adding GAD consistently reduces the FID discrepancy between the distilled student and the teacher model.}
\label{tab:fid_scores}
\begin{tabular}{llcc}
\toprule
\textbf{Model} & \textbf{Method} & \textbf{FID vs.\ COCO-30k} $\downarrow$ & \textbf{FID vs.\ Teacher} $\downarrow$ \\
\midrule
\multirow{3}{*}{\textbf{SD2}} & Teacher & 16.5505 & - \\
 & LADD & 16.7400 & 11.2920 \\
 & LADD + Ours & \textbf{16.5567} & \textbf{7.8463} \\
\midrule
\multirow{3}{*}{\textbf{PixArt-$\alpha$}} & Teacher & 28.4923 & - \\
 & TDM & \textbf{25.3909} & 5.8014 \\
 & TDM + Ours & 26.3235 & \textbf{5.1359} \\
\midrule
\multirow{3}{*}{\textbf{SANA}} & Teacher & 26.5129 & - \\
 & SiD & 28.6817 & 8.4830 \\
 & SiD + Ours & \textbf{27.2026} & \textbf{7.3596} \\
\bottomrule
\end{tabular}%
\end{table}

\hhy{
\section{2D Toy Example: Swiss Roll Visualization}
\label{sec:swiss_roll}
}

To intuitively visualize the ``geometric gap" and the structural degradation caused by standard pointwise distillation, we design a 2D toy experiment using the classic Swiss Roll dataset. The Swiss Roll represents a highly curved, low-dimensional data manifold embedded in a 2D space, making it an ideal testbed for evaluating whether a model preserves complex geometric structures.

We first train a standard diffusion model on the Swiss Roll dataset to serve as the Teacher model, which samples using 40 DDIM steps. Then, we distill this teacher into a 4-step Student model using a standard pointwise distillation objective, and compare it against a 4-step Student trained with our Geometry-Aware Distillation (GAD). 

\begin{figure}[ht]
    \centering
    \includegraphics[width=\linewidth]{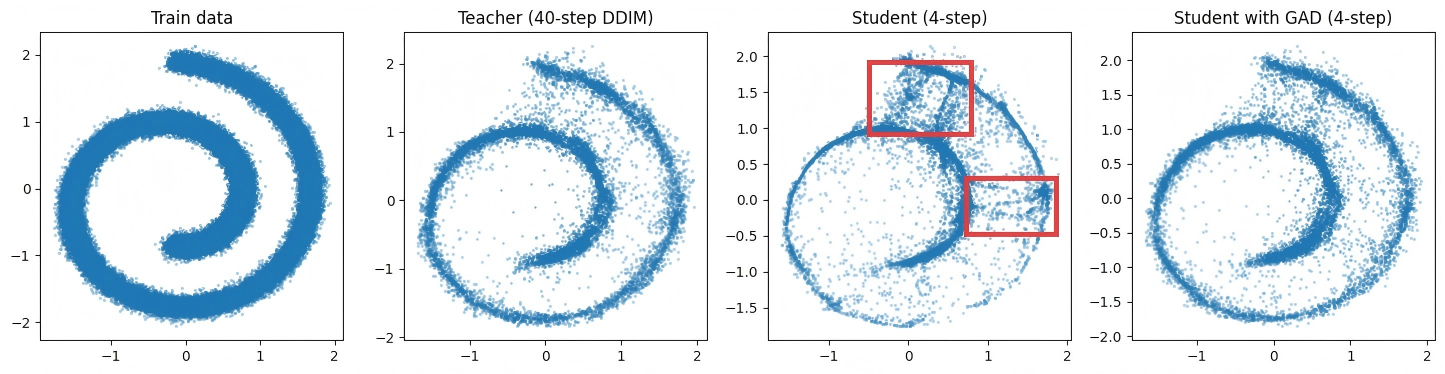} 
    \vspace{-0.2cm}
    \caption{\textbf{A Swiss Roll toy example visualizing the restoration of geometry.} Left to right: Ground truth training data, Teacher model (40-step DDIM), standard Student (4-step), and our GAD Student (4-step). Standard distillation leads to structural ``shortcuts" (red boxes) across the complex curves, causing severe distribution shifts. In contrast, GAD accurately preserves the teacher's original geometry and manifold curvature.}
    \label{fig:swiss_roll}
\end{figure}

As illustrated in Fig.~\ref{fig:swiss_roll}, the Teacher model successfully learns the highly curved data distribution. However, the standard Student model struggles to capture the continuous curvature. Because its pointwise objective optimizes for averaged responses, it attempts to map the distribution via erroneous ``shortcuts'' (highlighted by the red boxes). These shortcuts result in generated points falling into low-density regions that are entirely outside the original data manifold, causing significant distribution shifts. 

In contrast, the Student trained with GAD eliminates these structural shortcuts. By explicitly aligning the Jacobian-vector products (JVP), GAD forces the student to respect the local curvature and directional gradients of the teacher's mapping. Consequently, the GAD Student preserves the intricate geometry of the original manifold, yielding a distribution that is consistent with both the Teacher model and the ground-truth training data.

\section{More Visualization Results}
\label{app:visualize}

\begin{figure*}[t]
  \centering
  \includegraphics[width=\linewidth]{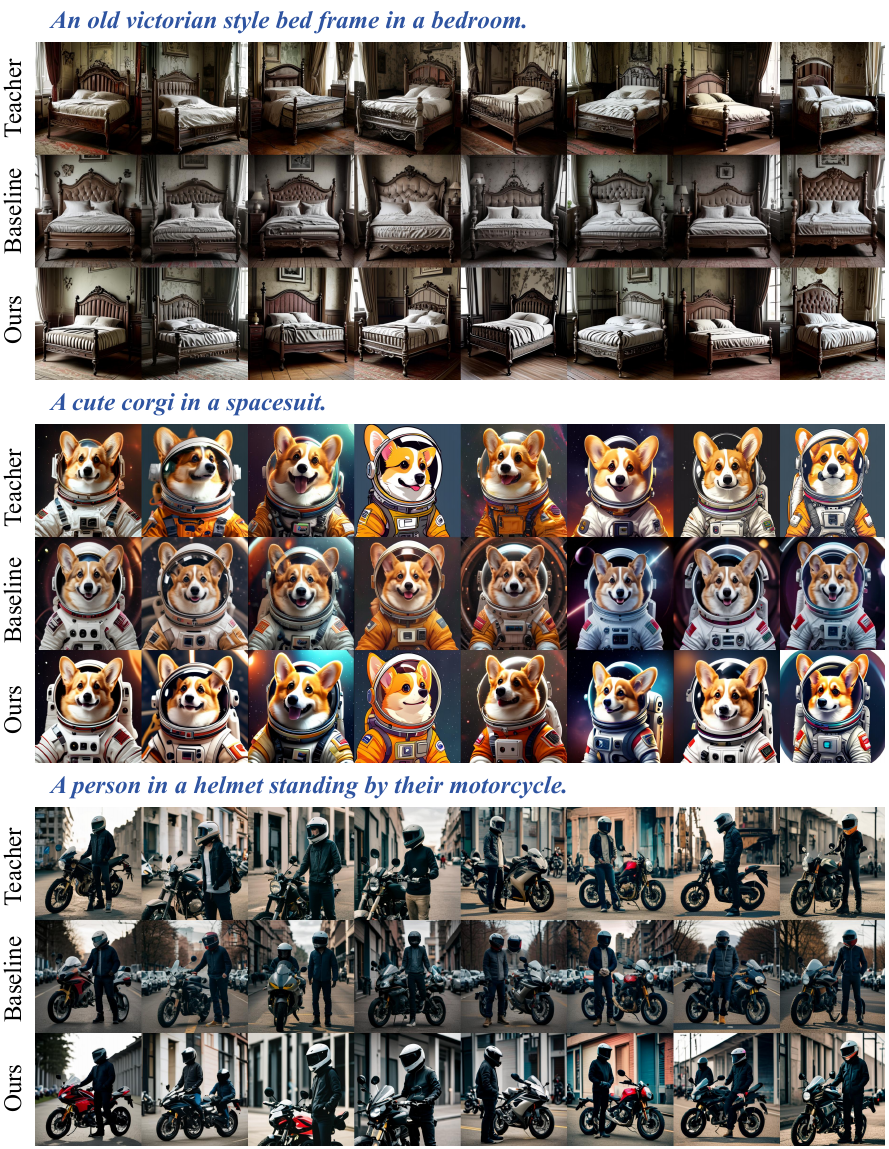}
  \caption{\textbf{More visualization of diversity improvement.} The experiments are conducted on the SANA model using the Score Identity Distillation (SiD) as the foundational distillation framework.
  }
  \label{fig:app-div}
\end{figure*}

\begin{figure*}[t]
  \centering
  \includegraphics[width=\linewidth]{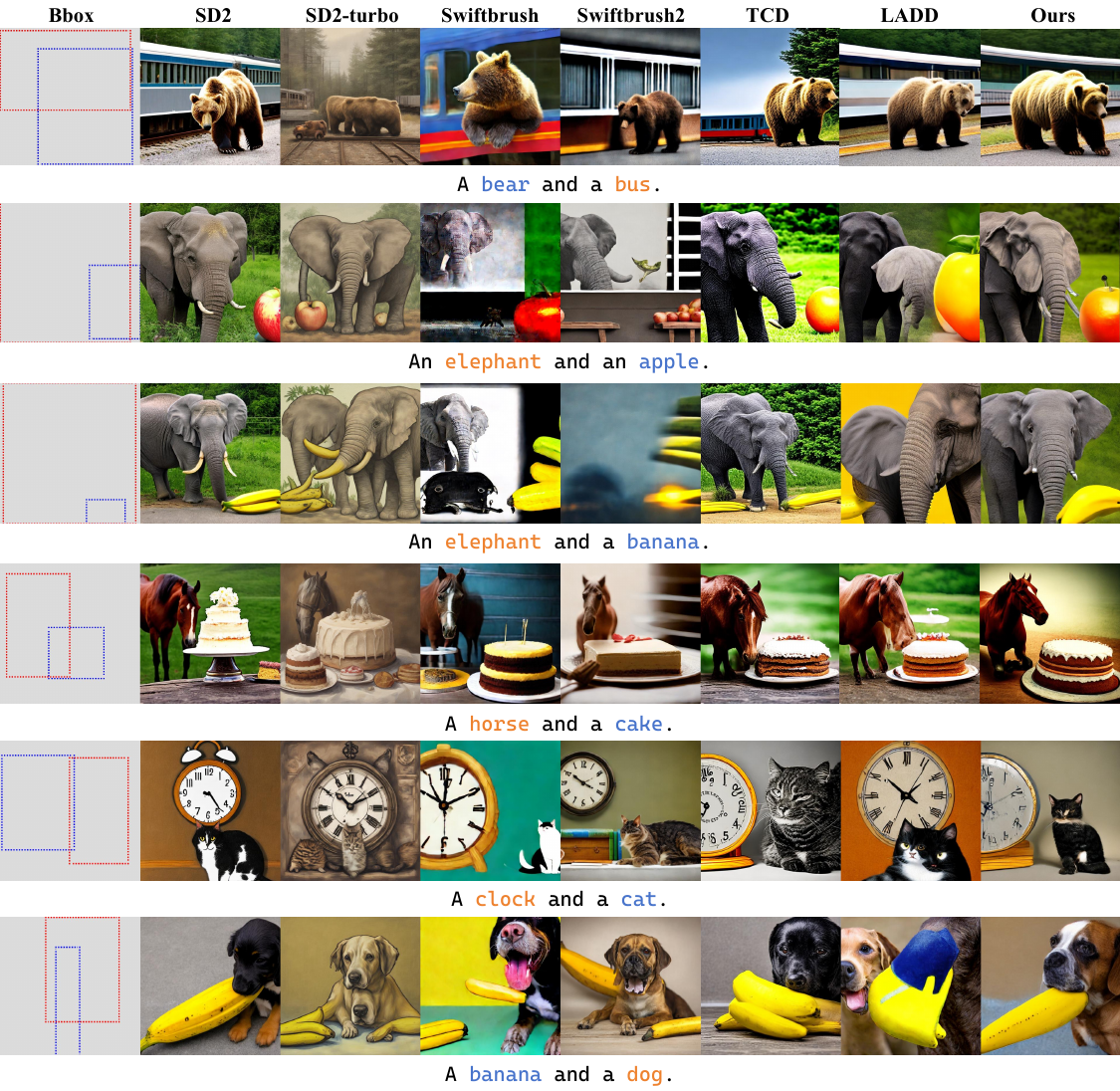}
  \caption{\textbf{More visualization of noise-based layout control.} The experiments are conducted on the Stable Diffusion v2 (SD2) model using LADD as the foundational distillation framework.
  }
  \label{fig:app-layout}
\end{figure*}

\end{document}